# Learning Hybrid Representation by Robust Dictionary Learning in Factorized Compressed Space

Jiahuan Ren, Zhao Zhang, *Senior Member, IEEE*, Sheng Li, *Senior Member, IEEE*, Yang Wang, Guangcan Liu, *Senior Member, IEEE*, Shuicheng Yan, *Fellow, IEEE*, and Meng Wang

*Abstract* — In this paper, we investigate the robust dictionary learning (DL) to discover the hybrid salient low-rank and sparse representation in a factorized compressed space. *A Joint Robust Factorization and Projective Dictionary Learning* (J-RFDL) model is presented. The setting of J-RFDL aims at improving the data representations by enhancing the robustness to outliers and noise in data, encoding the reconstruction error more accurately and obtaining hybrid salient coefficients with accurate reconstruction ability. Specifically, J-RFDL performs the robust representation by DL in a factorized compressed space to eliminate the negative effects of noise and outliers on the results, which can also make the DL process efficient. To make the encoding process robust to noise in data, J-RFDL clearly uses sparse $L_{2,1}$-norm that can potentially minimize the factorization and reconstruction errors jointly by forcing rows of the reconstruction errors to be zeros. To deliver salient coefficients with good structures to reconstruct given data well, J-RFDL imposes the joint low-rank and sparse constraints on the embedded coefficients with a synthesis dictionary. Based on the hybrid salient coefficients, we also extend J-RFDL for the joint classification and propose a discriminative J-RFDL model, which can improve the discriminating abilities of learnt coefficients by minimizing the classification error jointly. Extensive experiments on public datasets demonstrate that our formulations can deliver superior performance over other state-of-the-art methods.

## I. Introduction

WITH the increasing complexity of contents, diversity of distribution and high-dimensionality of real data, how to represent data efficiently for subsequent classification or clustering still remains an important research topic [1-3][9][50]. To represent data, some feasible methods can be used, such as sparse representation (SR) by dictionary learning (DL) [4-8], low-rank coding [9-10][15][38-39] and matrix factorization [11][12], which are inspired by the fact that high-dimensional data can usually be characterized by applying a low-dimensional or compressed space in which the possible noise and redundant information can be removed in addition to preserving the useful information and important structures.

Dictionary learning (DL) algorithms mainly study the topics on seeking an over-complete dictionary for representation and classification, which can be roughly divided into unsupervised and discriminative ones. Unsupervised DL methods do not use any supervised prior information (e.g., label information of the training data) and mainly aim at minimizing the reconstruction error to produce a reliable dictionary and a set of informative coefficients. To represent data, SR approximates each sample by a linear combination of a few items from a dictionary [4], so the quality of learnt dictionary is a dominating factor for SR. To obtain a reliable dictionary, many unsupervised DL methods [20][47-48] have been proposed, of which *K-Singular Value Decomposition* (KSVD) [20] is one most representative model and it seeks a dictionary from the whole training set. Note that for unsupervised learning by factorization, *Nonnegative Matrix Factorization* (NMF) [11] and *Concept Factorization* (CF) [12] are two most popular ones, and they both aim at factorizing the original data into the product between a set of nonnegative basis vectors and the compressed new representation. CF model is a variant of NMF by expressing each sample with a linear combination of cluster centers [12]. Low-rank coding is also an important representation model, such as [9][13][34]. Several representative feature embedding based unsupervised low-rank representation (LRR) models are *Inductive Robust Principal Component Analysis* (IRPCA) [14], *Latent LRR* (LatLRR) [15] and *Regularized LRR* (rLRR) [16] that can explicitly obtain a low-rank projection to extract salient low-rank features from given data and remove noise at the same time. More recently, an *Inductive Joint Low-Rank and Sparse Principal Feature Coding* (I-LSPFC) model [18] was presented for enhancing the representations by jointly exploiting the low-rank and sparse feature spaces. Note that unsupervised models mainly consider representing data, but cannot process the classification task.

In contrast to the unsupervised DL methods, discriminative approaches make full use of label information of samples when available to improve the joint representation and classification abilities [21-22][25-31][38][40]. Existing supervised methods can be further divided into two categories. The first category aims at learning the overall dictionary and the representative methods are *Discriminative KSVD* (D-KSVD) [25] and *Label Consistent KSVD* (LC-KSVD) [27]. D-KSVD obtains a desired dictionary of supporting the inter-class discrimination, while LC-KSVD learns a discriminative dictionary by adding a label consistent constraint and combining it with reconstruction error and classification error to form a unified model. The second

- J. Ren is with the School of Computer Science and Technology, Soochow University, Suzhou 215006, China (e-mails: hmzry10086@outlook.com)
- Z. Zhang is with the Key Laboratory of Knowledge Engineering with Big Data (Hefei University of Technology), Ministry of Education, China; also with Soochow University, Suzhou, China. (e-mail: cszzhang@gmail.com)
- Y. Wang and M. Wang are now with the Key Laboratory of Knowledge Engineering with Big Data (Ministry of Education), Hefei University of Technology, Hefei, China. (e-mails: yeungwangresearch@gmail.com, eric.mengwang@gmail.com)
- S. Li is with the Department of Computer Science, University of Georgia, Athens, GA 30602, USA (e-mail: sheng.li@uga.edu)
- G. Liu is with the School of Information and Control, Nanjing University of Information Science and Technology, Nanjing, China (e-mail: gcliu@nuist.edu.cn)
- S. Yan is with YITU Technology; also with the National University of Singapore, Singapore. (e-mail: shuicheng.yan@yitu-inc.com)

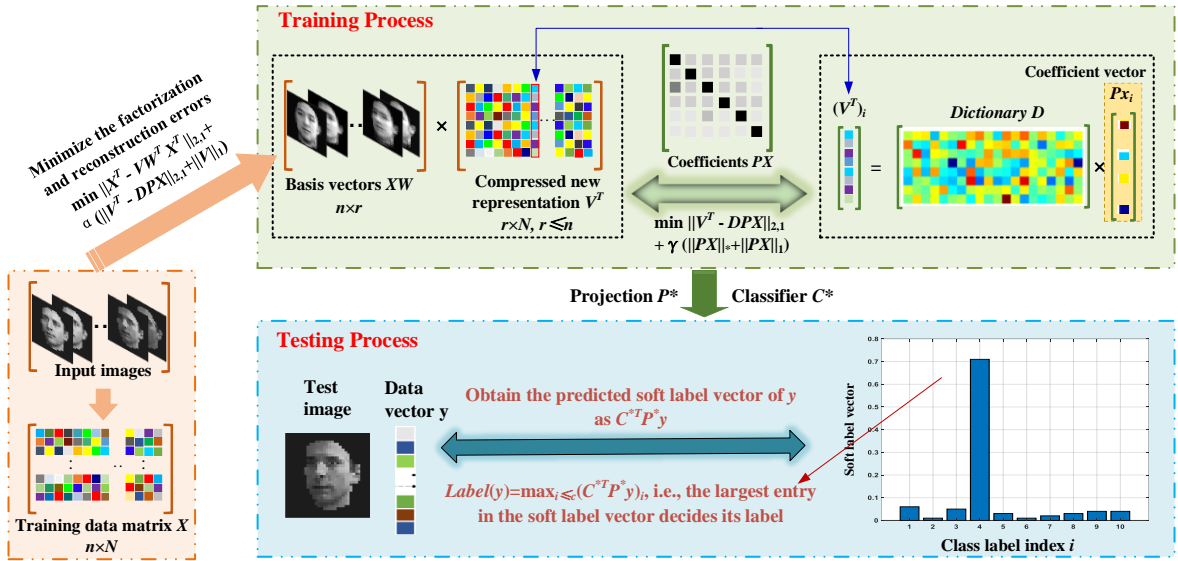

**Fig. 1:** The schematic diagram of our proposed J-RFDL and DJ-RFDL frameworks for training and testing.

category aims at learning the category-specific dictionaries to improve the discrimination by assigning each sub-dictionary to a single subject class, i.e., structured dictionary learning. In this category, popular methods are *Low-rank Shared DL* (LRSDL) [52], *Analysis Discriminative Dictionary Learning* (ADDL) [22], *Dictionary Learning with Structured Incoherence* (DLSI) [21] and *Projective Dictionary Pair Learning* (DPL) [17]. DPL aims to obtain a structured synthesis dictionary and an analysis dictionary jointly, and ADDL is based on the idea of DPL and it aims to compute a structured analysis discriminative dictionary and an analysis multiclass classifier.

It is worth noting that aforementioned existing methods still suffer from some shortcomings. First, most existing methods obtain the sparse representations and dictionaries in the original input space that usually consists of various noise, redundant information and even errors, so the data representation ability may be potentially degraded. Second, most existing methods encode the reconstruction error using the Frobenius-norm that is sensitive to noise and outliers, so the resulted reconstruction error may be inaccurate in reality. Thus, it will be helpful to present a robust DL approach that can work in a compressed feature subspace from which redundant information, noise and outliers are removed. Third, existing models usually compute the coefficients of each new sample by using a time-consuming reconstruction process with a well-trained dictionary, such as D-KSVD and LC-KSVD. Although DPL and ADDL aim to address this issue by calculating a synthesis dictionary jointly, they both did not consider regularizing the synthesis dictionary to obtain the salient low-rank and sparse coefficients. Most real data can usually be represented using a sparse and/or low-rank subspace due to the intrinsic low-dimensional characteristics [4-10]. Thus, without considering the joint sparse and low-rank constraints properly, the resulted structures of the coefficients may not represent the given data appropriately and accurately.

Motivated by the above existing shortcomings, in this paper we propose a new robust dictionary learning framework called *Joint Robust Factorization and Projective Dictionary Learning* (J-RFDL) for recovering the joint salient low-rank and sparse coefficients in a compressed factorized feature subspace. To be specific, J-RFDL integrates the robust factorization, projective DL and the embedded low-rank & sparse coding into a unified framework. We describe the major contributions as follows:

(1) J-RFDL performs the robust representation learning in a factorized compressed space in which unfavorable features and redundant information are removed from original data.

(2) To make the coding process robust to noise and outliers in data, and potentially reduce the factorization and reconstruction errors, J-RFDL uses the sparse and robust $L_{2,1}$-norm [35-37].

(3) J-RFDL can obtain the salient low-rank and sparse codes for accurate representations. Based on the salient coefficients, a *Discriminative J-RFDL* (DJ-RFDL) is also presented.

The paper is outlined as follows. In Section II, we review the related work briefly. In Sections III and IV, we present J-RFDL and DJ-RFDL, respectively. Section V shows the connections with other methods. Section VI shows the experimental results and analysis. Finally, the paper is concluded in Section VII.

## II. RELATED WORK

### A. Nonnegative Concept Factorization (CF)

Given a set of samples $X = [x_1, x_2 \cdots, x_N] \in \mathbb{R}^{n \times N}$, where $x_i \in \mathbb{R}^n$ is a sample that is represented by an *n*-dimensional vector, *n* is the original dimensionality and *N* is the number of samples. Note that the nonnegative factorization can be mathematically represented as finding two nonnegative matrix factors $U \in \mathbb{R}^{n \times r}$ and $V^T \in \mathbb{R}^{r \times N}$ whose product can best approximate *X*, i.e.,

$$X \approx UV^T, \text{ where } U \geq 0, V \geq 0,$$

where each column of *U* is the basis vector that captures the higher-level features of samples, and $V^T$ is low-dimensional representation of original data. From this viewpoint, the matrix factorization can be regarded as the dimensionality reduction method since it reduces the dimension from *n* to *r*. NMF and CF are the most popular factorization methods and the nonnegative constraints can learn the parts-based representation [11-12][42]. But how to perform NMF in reproducing kernel Hilbert space so that the powerful kernel method can be applied [12][42] is unclear. CF is proposed to solve this issue, and the advantage of

CF is that it can be applied to any representation space, which is implemented by representing each base $u_r$ by using a linear combination of samples, i.e., $u_r = \sum w_{jr} x_j$.

That is, CF aims to decompose $X$ and satisfy the equation $X \approx XWV^T$ by minimizing the following cost function:

$$\phi = \|X - XWV^T\|_F^2, \; s.t.\, W \geq 0, V \geq 0, \quad (1)$$

where $r$ is the rank of the factorization or the dimension of the reduced feature space, $V^T = [v_{ij}] \in \mathbb{R}^{r \times N}$ and $W = [w_{ij}] \in \mathbb{R}^{N \times r}$ with $w_{ij} \geq 0$ and $v_{ij} \geq 0$. CF applies an iteration strategy to update $V$ and $W$. Specifically, in the $(k+1)$-th iteration, $V$ and $W$ can be updated using the following multiplicative updating rules:

$$w_{jr}^{k+1} = w_{jr}^k \frac{(AV)_{jr}}{(AWV^TV)_{jr}}, \; v_{jr}^{k+1} = v_{jr}^k \frac{(AW)_{jr}}{(VW^TAW)_{jr}}, \quad (2)$$

where $A = X^T X$ is an auxiliary matrix.

*B. Dictionary Learning and Dictionary Pair Learning (DPL)*

Given the data matrix $X$, the standard DL algorithms solve the following general framework to compute the reconstructive dictionary $D$ and sparse representation $S$ of $X$:

$$\langle D, S \rangle = \arg\min_{D,S} \|X - DS\|_F^2 + \kappa \|S\|_p, \; s.t.\, \|d_i\|_2^2 \leq 1, \quad (3)$$

where $\kappa > 0$ denotes a scalar constant, $D = [d_1, \cdots d_K] \in \mathbb{R}^{n \times K}$ is an overall dictionary obtained by minimizing the reconstruction error $\|X - DS\|_F^2$. $\|S\|_p$ is the $L_p$-norm regularized term, where $S = [s_1 \cdots s_N] \in \mathbb{R}^{K \times N}$ is the coding coefficients of $X$ and usually the parameter $p$ is usually set to 0 or 1 to make the learned coefficients sparse. However, the computations of $L_0$-norm and $L_1$-norm are usually time-consuming in reality.

Suppose that label information of training set is known, i.e., the sample set can be expressed as $X = [X_1, \cdots X_l, \cdots, X_N] \in \mathbb{R}^{n \times N}$, where $X_l \in \mathbb{R}^{n \times N_l}$ is the training set of class $l$ and $N_l$ is the number of samples in the class $l$. To avoid the costly computation of $S$, DPL learns a synthesis dictionary $D$ and an analysis dictionary $P$ to approximate the coding coefficients without using costly $L_0$-norm/$L_1$-norm constraint by solving the following problem:

$$\langle P, D \rangle = \arg\min_{P,D} \sum_{i=1}^c \|X_i - D_i P_i X_i\|_F^2 + \lambda \|P_i \overline{X}_i\|_F^2, \; s.t.\, \|d_j\|_2^2 \leq 1, \quad (4)$$

where $\overline{X}_i$ is the complementary data matrix of $X_i$ in $X$. $D_i$ is a structured synthesis sub-dictionary of class $i$, $P_i$ is a structured analysis sub-dictionary of class $i$, $D_i \in \mathbb{R}^{n \times K}, P_i \in \mathbb{R}^{K \times n}$ and $K$ is the dictionary size. Clearly, by setting $P_i X_q \approx 0, \forall q \neq i$, the coefficient matrix $PX$ will be nearly block diagonal. $d_j$ is the $j$-th atom of dictionary $D$, and the constraint $\|d_j\|_2^2 \leq 1$ on $d_j$ is to avoid the trivial solution of $P_i = 0$ and make the algorithm stable.

## III. JOINT ROBUST FACTORIZATION AND PROJECTIVE DICTIONARY LEARNING FRAMEWORK (J-RFDL)

*A. The Objective Function*

We present the problem formulation of our J-RFDL that aims at learning the hybrid salient low-rank and sparse representations by robust dictionary learning in factorized compressed space. Specifically, it improves the data representations by enhancing the robustness to outliers and noise in data during both DL and matrix factorization processes, encoding the factorization and reconstruction errors accurately with row sparse constraint and delivering the hybrid salient representation coefficients with superior reconstruction power. To define the problem, J-RFDL integrates the robust factorization, robust dictionary learning and embedded low-rank & sparse representation learning into a unified model. By combining the factorization error $f(W, V^T)$ with the sample reconstruction error $g(D, P)$, we can define the following unified problem for our J-RFDL:

$$\min_{W,V,D,P} f(W, V^T) + g(D, P), \; s.t.\, W, V \geq 0, e^T D = e^T, \quad (5)$$

where $W$ is also a nonnegative matrix as in CF, $V^T$ is learnt new representation, $D$ is a dictionary and $P$ is a projection matrix to obtain the embedded low-rank & sparse coefficients $PX$. Clearly, $g(D, P)$ is the reconstruction error by robust dictionary learning over $V^T$. So, performing DL in the reduced feature space can be more efficient, especially when the dimension of $X$ is high. The sum-to-one constraint $e^T D = e^T$ can also prevent the big values of $D$ and make the problem stable, where $e$ is a vector of ones. $f(W, V^T)$ and $g(D, P)$ will be introduced shortly. The above problem can be alternated between the following steps:

*(1) Robust dictionary learning in the factorized compressed space for hybrid low-rank and sparse representation*

We first fix the new representation $V^T$ by robust sparse concept factorization to update the dictionary $D$ and hybrid low-rank & sparse coding coefficients $PX$ from the following problem:

$$\min_{D,P} g(D, P) = \alpha \|V^T - DPX\|_{2,1} + \gamma (\|PX\|_* + \|PX\|_1), s.t.\, e^T D = e^T, \quad (6)$$

where $\|V^T - DPX\|_{2,1}$ is the $L_{2,1}$-norm based reconstruction error term for robust DL. $L_{2,1}$-norm can minimize the reconstruction error as much as possible because it tends to make rows of the reconstruction error $V^T - DPX$ to be zeros [35]. In other words, minimizing the reconstruction error can enable $DPX$ to best approximate $V^T$. Besides, the $L_{2,1}$-norm based reconstruction is so robust to noise and outliers [35]. To make the coefficients $PX$ satisfy the hybrid low-rank and sparse properties to enhance the representation ability, the Nuclear-norm and $L_1$-norm are jointly regularized on $PX$, i.e., $\|PX\|_* + \|PX\|_1$, which is motivated by the LSPFC algorithm [18]. But note that J-RFDL differs from LSPFC in two aspects. First, LSPFC is a low-rank coding model for decomposing given data into a salient feature part and a sparse error part, while J-RFDL is a robust dictionary learning method. Second, LSPFC encodes salient low-rank and sparse features, while J-RFDL obtains the hybrid low-rank and sparse coefficients for the data representations.

*(2) Robust sparse concept factorization for compression:*

After the $D$ and projection $P$ are obtained, we can learn the new sparse representation $V^T$ by solving the following $L_{2,1}$-norm and $L_1$-norm regularized robust concept factorization problem:

$$\min_{W,V} f(W, V^T) = \|X^T - VW^T X^T\|_{2,1} + \alpha (\|V^T - DPX\|_{2,1} + \|V\|_1), \quad (7)$$

$s.t.\, W, V \geq 0$, where $\|X^T - VW^T X^T\|_{2,1}$ is the $L_{2,1}$-norm based factorization error, $\|V\|_1$ is the $L_1$-norm constraint so that learnt $V^T$ holds the sparse properties, and the product of $X$, $W$ and $V^T$ is an approximation to $X$. The $L_{2,1}$-norm can also minimize the error as much as possible by forcing many rows of the error to be zeros [35]. It is clear that $V^T$ is obtained via minimizing the joint reconstruction error $\|X^T - VW^T X^T\|_{2,1} + \alpha \|V^T - DPX\|_{2,1}$, i.e., the robust projective DL process also contributes to improving the representation ability of $V^T$. In other words, the optimization of $V^T$ is also associated with the pre-calculated dictionary $D$ and hybrid coefficients $PX$ in the last step.

By combining the equations in Eqs.(5), (6) and (7), the final objective function of our J-RFDL is formulated as

$$\min_{D,P,W,V} \|X^T - VW^T X^T\|_{2,1} + \alpha \left(\|V^T - DPX\|_{2,1} + \|V\|_1\right) + \gamma \left(\|PX\|_* + \|PX\|_1\right),$$
$$s.t. \ W, V \geq 0, e^T D = e^T \qquad (8)$$

where the optimal $P^*X$ is also regarded the "hybrid lowest-rank and sparsest representation" of the original data. The schematic diagram of our J-RFDL framework is shown in Fig.1, where we present the training and testing phases. Next, we describe the optimization procedures of our J-RFDL.

### B. Optimization

The variables in the objective function of J-RFDL depend on each other, so we follow the common procedures to solve it by an alternate strategy and use the inexact Augmented Lagrange Multiplier (ALM) method [19] for efficiency. We first convert the objective function into the following equivalent one:

$$\min_{\substack{D,P,W,V \\ J,S,F}} \|X^T - VW^T X^T\|_{2,1} + \alpha \left(\|V^T - DPX\|_{2,1} + \|F\|_1\right) + \gamma \left(\|J\|_* + \|S\|_1\right),$$
$$s.t. \ V = F, PX = J, PX = S, W \geq 0, V \geq 0, e^T D = e^T \qquad (9)$$

from which the Lagrange function $\wp$ can be constructed as

$$\wp = \|X^T - VW^T X^T\|_{2,1} + \alpha \left(\|V^T - DPX\|_{2,1} + \|F\|_1\right) + \gamma \left(\|J\|_* + \|S\|_1\right)$$
$$+ \langle Y_1, V - F\rangle + \langle Y_2, PX - J\rangle + \langle Y_3, PX - S\rangle \qquad ,(10)$$
$$+ \frac{\mu}{2}\left(\|V - F\|_F^2 + \|PX - J\|_F^2 + \|PX - S\|_F^2\right)$$

with respect to $W, V \geq 0, e^T D = e^T$, where $Y_1, Y_2, Y_3$ are Lagrange multipliers, and $\mu$ is a positive weighting parameter. By using the inexact ALM, our J-RFDL updates the variables by

$$\left(D_{k+1}, P_{k+1}, W_{k+1}, V_{k+1}, J_{k+1}, S_{k+1}, F_{k+1}\right) = \arg\min_{D,P,W,V,J,S,F} \wp_k$$
$$Y_1^{k+1} = Y_1^k + \mu(V_k - F_k), Y_2^{k+1} = Y_2^k + \mu(P_k X - J_k) \qquad . \quad (11)$$
$$Y_3^{k+1} = Y_3^k + \mu(P_k X - S_k), \mu_{k+1} = \min(\eta\mu_k, \max_\mu)$$

Then, the optimization of J-RFDL can be detailed as follows.

**1) Fix others, update $J, S, F$:**
We first update the auxiliary variable $J$. By removing the terms irrelevant to $J$, we can update $J_{k+1}$ at the $(k+1)$-th iteration as

$$J_{k+1} = \arg\min_J \frac{\gamma}{\mu_k} \|J\|_* + \frac{1}{2}\left\|J - \left(P_k X + Y_2^k/\mu_k\right)\right\|_F^2. \qquad (12)$$

Following the procedure for solving the Nuclear-norm based problem [19], $J$ can be updated by SVD. More specifically, let $\mho_\varepsilon[x] = \text{sgn}(x)\max(|x| - \varepsilon, 0)$ be the shrinkage operator, we can compute $J_{k+1}$ by the singular value thresholding algorithm as $J_{K+1} = M_J \mho_{\gamma/\mu}[\Sigma_J] Q_J^T$, where $M_J \Sigma_J Q_J^T$ represents the SVD of $P_k X + Y_2^k/\mu_k$. Note that the updating rules of both $S$ and $F$ are similarly as solving $J$ by the scalar shrinkage operator [19] as

$$S_{k+1} = \arg\min_S \frac{\gamma}{\mu_k} \|S\|_1 + \frac{1}{2}\left\|S - \left(P_k X + Y_3^k/\mu_k\right)\right\|_F^2, \qquad (13)$$

$$F_{k+1} = \arg\min_F \frac{\alpha}{\mu_k} \|F\|_1 + \frac{1}{2}\left\|F - \left(V_k + Y_1^k/\mu_k\right)\right\|_F^2. \qquad (14)$$

**2) Fix others, update $D, P$:**
With the other variables fixed, we can easily both $D$ and $P$ by performing robust dictionary learning. By removing the terms that are irrelevant to $D$ and $P$ from $\wp$, we have

$$\min_{D,P} \alpha \|V^T - DPX\|_{2,1} + \langle Y_2, PX - J\rangle + \langle Y_3, PX - S\rangle$$
$$+ \frac{\mu}{2}\left(\|PX - J\|_F^2 + \|PX - S\|_F^2\right), s.t. \ e^T D = e^T \qquad . \quad (15)$$

Based on the definition of $L_{2,1}$-norm [35], we can easily have $\|V^T - DPX\|_{2,1} = 2tr\left((V^T - DPX)^T Q(V^T - DPX)\right)$, where $Q$ denotes a diagonal matrix with entries $q_{ii} = 1/(2\|\chi^i\|_2)$ and $\chi^i$ is $i$-th row vector of matrix $V^T - DPX$. Suppose that each vector $\chi^i \neq 0$, the above problem can be approximated as

$$\min_{D,P,Q} tr\left(\alpha(V^T - DPX)^T Q(V^T - DPX)\right) + \langle Y_2, PX - J\rangle$$
$$+ \langle Y_3, PX - S\rangle + \frac{\mu}{2}\left(\|PX - J\|_F^2 + \|PX - S\|_F^2\right) \qquad . \quad (16)$$

By taking the derivative of the above equation w.r.t. $D$ and zeroing the derivative, we can update the dictionary $D_{k+1}$ as

$$D_{k+1} = Q_k^{-1}\left(Q_k V_k^T X^T P_k^T\right)\left(P_k X X^T P_k^T\right)^{-1}, \ e^T D_{k+1} = e^T. \qquad (17)$$

Similarly, by taking the derivative w.r.t. $P$ and setting it to zero, we can update the projection $P_{k+1}$ as

$$P_{k+1} = \left(2\alpha D_{k+1}^T Q_k D_{k+1} + 2\mu_k I\right)^{-1} L_k \left(XX^T + \tau I\right)^{-1}. \qquad (18)$$

where $L_k = \left(2\alpha(XV_k Q_k D_{k+1})^T - Y_2^k X^T - Y_3^k X^T + \mu_k J_k X^T + \mu_k S_k X^T\right)$, and $\tau I$ with a small $\tau$ is to avoid the singularity issue.

After both $D$ and $P$ are obtained, we can update the entries of the diagonal matrix $Q$ as $(q_{ii})_{k+1} = 1/(2\|\chi_{k+1}^i\|_2)$, where $\chi_{k+1}^i$ is $i$-th row of matrix $V_{k+1}^T - D_{k+1} P_{k+1} X$.

**3) Fix others, update $W$ and $V$:**
We update $W$ and $V$ by performing the robust sparse concept factorization. After removing the irrelevant terms, we have

$$\min_{W,V} \wp(W,V) = \|X^T - VW^T X^T\|_{2,1} + \alpha \|V^T - DPX\|_{2,1}$$
$$+ \langle Y_1, V - F\rangle + \frac{\mu}{2}\|V - F\|_F^2, \ s.t. \ W, V \geq 0 \qquad . \quad (19)$$

For the optimization of $W$ and $V$, we employ the iterative multiplicative updating rules to obtain the local optima. By the matrix expression and the definition of $L_{2,1}$-norm [35], we first transform Eq.(19) into the following equivalent one:

$$\min_{W,V,Q,G} \wp(W,V,Q,G) = tr\left((X^T - VW^T X^T)^T G(X^T - VW^T X^T)\right)$$
$$+ tr\left(\alpha(V^T - DPX)^T Q(V^T - DPX)\right) \qquad , \quad (20)$$
$$+ tr\left(Y_1^T(V - F)\right) + \frac{\mu}{2} tr\left((V - F)^T(V - F)\right)$$

where $G$ is a diagonal matrix with entries $g_{ii} = 1/(2\|\psi^i\|_2)$ and $\psi^i$ is $i$-th row vector of the matrix $X^T - VW^T X^T$. Note that the above problem holds if each $\psi^i \neq 0$ and $\psi^i \neq 0$. Let $\tau_{i,j}$ and $\varsigma_{i,j}$

**Algorithm 1**: Optimization procedures of J-RFDL

**Inputs:** Training set $X$, dictionary size $K$, parameters $\alpha, \gamma$.
**Initialization:** $J_k = 0$, $S_k = 0$, $F_k = V_k = 0$, $E_k = 0$, $\varepsilon = 10^{-7}$, $Y_1^k = 0$, $Y_2^k = 0$, $Y_3^k = 0$, $\mu = 10^{-6}$, $\max_\mu = 10^6$, $\eta = 1.12$, $k = 0$; Initialize $Q = I$ and $G = I$, and initialize $D_k$, $P_k$, $W_k$ and $V_k$ as random matrices.
**While** *not converged* **do**
1. Fix others and update the low-rank matrix $J$ by Eq.(12);
2. Fix others and update the sparse matrix $S$ by Eq.(13);
3. Fix others and update the sparse matrix $F$ by Eq.(14);
4. Fix others and update the dictionary $D$ by Eq.(17);
5. Fix others and update the projection $P$ by Eq.(18);
6. Fix others and update the basis vectors $W$ by Eq.(26);
7. Fix others and update the new representation $V$ by Eq.(27);
8. Update the entries of the diagonal matrices $Q$ and $G$ as $(q_{ii})_{k+1} = 1/(2\|\chi_{k+1}^i\|_2)$ and $(g_{ii})_{k+1} = 1/(2\|\psi_{k+1}^i\|_2)$.
9. Update the multipliers $Y_1, Y_2, Y_3$ by Eq.(11);
10. Update the parameter $\mu$ with $\mu_{k+1} = \min(\eta\mu_k, \max_\mu)$;
11. Convergence check: suppose $\max(\|PX - J\|_\infty, \|PX - S\|_\infty, \|V - F\|_\infty) \leq \varepsilon$, stop; else $k = k+1$.

**End while**
**Outputs:** $D^* \leftarrow D_{k+1}$, $P^* \leftarrow P_{k+1}$.

be the Lagrange multipliers for constraints $w_{i,j} \geq 0$ and $v_{i,j} \geq 0$. By denoting $\tau = [\tau_{i,j}]$ and $\varsigma = [\varsigma_{i,j}]$, the Lagrange function $\Im$ of the above problem can be obtained as

$$\Im = tr\left((X^T - VW^T X^T)^T G(X^T - VW^T X^T)\right)$$
$$+ tr\left(\alpha(V^T - DPX)^T Q(V^T - DPX)\right) \quad . \quad (21)$$
$$+ tr(Y_1^T(V-F)) + \frac{\mu}{2}tr((V-F)^T(V-F)) + tr(\tau W^T) + tr(\varsigma V^T)$$

By taking the partial derivatives *w.r.t.* $W$ and $V$, we have

$$\partial\Im/\partial W = -2X^T XGV + 2X^T XWV^T GV + \tau, \quad (22)$$

$$\partial\Im/\partial V = -2GX^T XW + 2GVW^T X^T XW$$
$$+ \alpha(2VQ - 2X^T P^T D^T Q) + Y_1 + \mu(V-F) + \varsigma \quad . \quad (23)$$

By using the Kuhn–Tucker conditions $\tau_{i,j} w_{i,j} = 0$, $\varsigma_{i,j} v_{i,j} = 0$, we can obtain the following equations:

$$(X^T XWV^T QV)_{i,j} w_{i,j} - (X^T XQV)_{i,j} w_{i,j} = 0, \quad (24)$$

$$(2GVW^T X^T XW + 2\alpha VQ + Y_1 + \mu V)_{i,j} v_{i,j} -$$
$$(2GX^T XW + 2\alpha X^T P^T D^T Q + \mu F)_{i,j} v_{i,j} = 0 \quad . \quad (25)$$

These equations lead to the following updating rules:

$$w_{i,j} \leftarrow w_{i,j} \frac{(X^T XGV)_{i,j}}{(X^T XWV^T GV)_{i,j}}, \quad (26)$$

$$v_{i,j} \leftarrow v_{i,j} \frac{(2GX^T XW + 2\alpha X^T P^T D^T Q + \mu F)_{i,j}}{(2GVW^T X^T XW + 2\alpha VQ + Y_1 + \mu V)_{i,j}}. \quad (27)$$

After both $W$ and $V$ are obtained, we can update the entries of the diagonal matrix $G$ as $(g_{ii})_{k+1} = 1/(2\|\psi_{k+1}^i\|_2)$, where $\psi_{k+1}^i$ is $i$-th row of $X_{k+1}^T - V_{k+1} W_{k+1}^T X^T$. For complete presentation of the optimization procedures of our J-RFDL, we summarize them in Algorithm 1, where diagonal matrices $Q$ and $G$ are initialized to be the identity matrices as suggested in [24][35] that has shown that this kind of regularization can generally perform well. Note that an early version of this work has been presented in [59]. This paper also presents a discriminant extension, provides the detailed analysis of the formulation, and conducts a thorough evaluation on the tasks of representation and classification.

*C. Computational Time Complexity*

In the problem of our J-RFDL, the variables $D, P, W$ and $V$ are updated alternately. In each iteration, the time complexities of updating $D, P, W, V$ can be calculated. We employ the big $O$ notation to describe the time complexity as [46]. For the robust factorization error $f(W, V^T)$ and the rules in Eqs.(26-27), our J-RFDL has the same computational time complexity as CF for updating $W$ and $V$, i.e., $O(N^2 r)$, where $N$ is the number of samples in $X$ and $r$ is the dimension of the new representation $V$. For the reconstruction error $g(D, P)$, the time complexities of updating $P$ and $D$ are $O(K^3 + nNr + n^3)$ and $O(r^3 + rNn + K^3)$ in each iteration. Thus, $O(k(K^3 + nNr + n^3))$ is the overall cost of our J-RFDL, when the updates stop after $k$ iterations.

## IV. DISCRIMINATIVE J-RFDL (DJ-RFDL) FRAMEWORK

Because J-RFDL focuses on the unsupervised representation [15-16][32-33] and has to seek an extra classifier by using the coefficients, but this cannot minimize the representation and classification errors jointly. To address this issue, we extend J-RFDL for joint classification by combining the classification error based on the hybrid salient coefficients and derive a more general model termed *Discriminant J-RFDL* (DJ-RFDL).

*A. Problem Formulation of DJ-RFDL*

When label information $\Gamma = [l(x_1), l(x_2), \cdots l(x_N)]$ of training set $X$ is available, where $l(x_i)$ is the label of each training data $x_i$, we can include a $L_{2,1}$-norm based regressive classification error $\|H^T - X^T P^T C\|_{2,1}$ based on the salient coefficients $PX$ into the problem of J-RFDL for the simultaneous minimization, where $C \in \mathbb{R}^{K \times c}$ is a predictive classifier, $c$ is the number of classes and $C^T P x_i$ is the estimated soft label vector of each $x_i$ by $C$. That is, DJ-RFDL integrates the robust hybrid salient representation via dictionary learning and robust classifier learning. Note that this operation can ensure the hybrid $PX$ to be joint optimal to predict the labels of samples. The objective function of DJ-RFDL is

$$\min_{D,P,W,V,C} \|X^T - VW^T X^T\|_{2,1} + \alpha(\|V^T - DPX\|_{2,1} + \|V\|_1)$$
$$+ \gamma(\|PX\|_* + \|PX\|_1) + \beta(\|H^T - X^T P^T C\|_{2,1} + \|C\|_F^2), \quad (28)$$
$$s.t. \; W, V \geq 0$$

where $H = [h_1, h_2, \ldots h_N] \in \mathbb{R}^{c \times N}$ is a pre-defined label matrix of all training samples based on the labels $\Gamma = [l(x_1), l(x_2), \cdots l(x_N)]$, $h_j$ is the initial label vector of sample $x_j$, $h_{i,j}$ is the $i$-th entry of the column vector $h_j$ and $h_{i,j} = 1$ if $x_j$ belongs to the class $i$ [18], $1 \leq i \leq c$. The $L_{2,1}$-norm on classification error can enable the metric to be robust to outliers and noise [24][35][37], and the use of $L_{2,1}$-norm on the classification error essentially assumes that some classes have larger classification errors than others. The classification error can be reduced, since $L_{2,1}$-norm can also make many rows of the classification error shrink to zeros [24][35]. Clearly, DJ-RFDL is more general than J-RFDL. Next, we describe its optimization procedures.

## B. Optimization

Due to the convexity of the problem of DJ-RFDL, we also use the inexact ALM method [19] to optimize its objective function. Let $E = H^T - X^T P^T C$ be the classification error, we can convert Eq.(28) into the following equivalent one:

$$\min_{\substack{D,P,W,V,\\J,S,F,C,E}} \|X^T - VW^T X^T\|_{2,1} + \alpha\left(\|V^T - DPX\|_{2,1} + \|F\|_1\right)$$
$$+ \gamma\left(\|J\|_* + \|S\|_1\right) + \beta\left(\|E\|_{2,1} + \|C\|_F^2\right) \quad (29)$$
$$s.t.\ V = F,\ PX = J,\ PX = S,\ H^T = X^T P^T C + E,\ W \geq 0,\ V \geq 0$$

The augmented Lagrange function can be similarly defined as

$$\Omega = \wp + \beta\left(\|E\|_{2,1} + \|C\|_F^2\right) + \langle Y_4, H^T - X^T P^T C - E\rangle$$
$$+ \frac{\mu}{2}\left(\|H^T - X^T P^T C - E\|_F^2\right),\ s.t.\ W, V \geq 0 \quad (30)$$

where $\wp$ is the Lagrange function of J-RFDL. By the inexact ALM, DJ-RFDL updates the variables alternately as:

$$(D_{k+1}, P_{k+1}, W_{k+1}, V_{k+1}, C_{k+1}, J_{k+1}, S_{k+1}, F_{k+1}, E_{k+1}) = \arg\min_{D,P,W,V,C,J,S,F} \Omega$$
$$Y_1^{k+1} = Y_1^k + \mu(V_k - F_k),\ Y_2^{k+1} = Y_2^k + \mu(P_k X - J_k) \quad (31)$$
$$Y_3^{k+1} = Y_3^k + \mu(P_k X - S_k),\ Y_4^{k+1} = Y_4^k + \mu(H^T - X^T P_k^T C_k)$$

Then, the optimization of DJ-RFDL can be described as:

### 1) Fix others, update J, S, F, W, V, D respectively:

The updating rules of J, S, F, W, V, D are the same as J-RFDL.

### 2) Fix others, update the projection P and Q:

By removing the terms irrelevant to P and Q from $\Omega$, we have

$$\min_{P,Q} \Omega(P,Q) = tr\left(\alpha(V^T - DPX)^T Q(V^T - DPX)\right) + \langle Y_2, PX - J\rangle$$
$$+ \langle Y_3, PX - S\rangle + \langle Y_4, H^T - X^T P^T C - E\rangle \quad (32)$$
$$+ \frac{\mu}{2}\left(\|PX - J\|_F^2 + \|PX - S\|_F^2 + \|H^T - X^T P^T C - E\|_F^2\right)$$

By taking the derivative of $\Omega(P,Q)$ w.r.t. P and setting it to zero, we can update the underlying projection $P_{k+1}$ as

$$P_{k+1} = \left(2\alpha D_{k+1}^T Q_k D_{k+1} + 2\mu_{k+1}I + \mu C_k C_k^T\right)^{-1} Z_k (XX^T + \tau I)^{-1}. \quad (33)$$

where $R_k = 2\alpha(XV_{k+1} Q_k D_{k+1})^T$, $T_k = C_k Y_4^T X^T + \mu_k C_k H X^T - \mu_k C_k E^T X^T$ and $Z_k = (R_k - Y_2^k X^T - Y_3^k X^T + \mu_k J_k X^T + \mu_k S_k X^T + T_k)$. After V, D and P are obtained, we can update Q by $(q_{ii})_{k+1} = 1/(2\|\chi_{k+1}^i\|_2)$, where $\chi_{k+1}^i$ is i-th row of the matrix $V_{k+1}^T - D_{k+1} P_{k+1} X$.

### 3) Fix others, update the classifier C and error E:

We first describe how to update C from the reduced problem:

$$\min_C \beta\|C\|_F^2 + \langle Y_4, H^T - X^T P^T C - E\rangle + \frac{\mu}{2}\|H^T - X^T P^T C - E\|_F^2. \quad (34)$$

By taking the derivative w.r.t. C and setting it to zero, we can update the classifier $C_{k+1}$ as

$$\nabla_{k+1} = P_{k+1} X Y_4^k / \mu_k + P_{k+1} X H^T - P_{k+1} X E_k$$
$$C_{k+1} = \left(P_{k+1} XX^T P_{k+1}^T + 2\beta I/\mu_k\right)^{-1} \nabla_{k+1} \quad (35)$$

Then, the error matrix E can be analogously inferred as

$$E_{k+1} = \arg\min_E \frac{\beta}{\mu_k}\|E\|_{2,1} + \frac{1}{2}\left\|E - \left(H^T - X^T P_{k+1}^T C_{k+1} + Y_4^k/\mu_k\right)\right\|_F^2. \quad (36)$$

The iterate $E_{k+1}$ can be obtained via the shrinkage operator as $E_{k+1} = \mho_{\beta/\mu}[\Sigma_E]$, where $\Sigma_E = (H^T - X^T P_{k+1}^T C_{k+1} + Y_4^k/\mu_k)$. For complete presentation of the optimization of our DJ-RFDL, we summarize the procedures in Algorithm 2.

---

**Algorithm 2**: Optimization procedures of DJ-RFDL

**Inputs:** Training data $X = \{x_i\}_{i=1}^N$, class label set $\Gamma = \{l(x_i)\}_{i=1}^N$ of X, dictionary size K, parameters $\alpha$, $\beta$ and $\gamma$.

**Initialization:** $J_k = 0$, $S_k = 0$, $F_k = V_k = 0$, $E_k = 0$, $\varepsilon = 10^{-7}$, $Y_1^k = 0$, $Y_2^k = 0$, $Y_3^k = 0$, $Y_4^k = 0$, $\mu = 10^{-6}$, $\max_\mu = 10^6$, $\eta = 1.12$; Initialize $Q = I$ and $G = I$, and initialize $D_k$, $P_k$, $W_k$ and $V_k$ as random matrices; Construct the initial label set $H = [h_1, h_2, \ldots h_N] \in \mathbb{R}^{c\times N}$; $k = 0$.

**While** *not converged* **do**
1. The optimization procedures of S, F, J, D, W and V are the same as those of J-RFDL in Algorithm 1;
2. Fix others and update the projection P by Eq.(33);
3. Update the entries of the diagonal matrices Q and G as $(q_{ii})_{k+1} = 1/(2\|\chi_{k+1}^i\|_2)$ and $(g_{ii})_{k+1} = 1/(2\|\psi_{k+1}^i\|_2)$.
4. Fix others and update the classifier C by Eq.(35);
5. Fix others and update the regression error E by Eq.(36);
6. Update the multipliers $Y_1, Y_2, Y_3$ and $Y_4$ by Eq.(31);
7. Update the parameter $\mu$ with $\mu_{k+1} = \min(\eta\mu_k, \max_\mu)$
8. Convergence check: if $\max(\|PX - J\|_\infty, \|PX - S\|_\infty, \|V - F\|_\infty, \|H^T - X^T P^T C - E\|_\infty) \leq \varepsilon$, stop; else $k = k+1$.

**End while**

**Outputs:** $D^* \leftarrow D_{k+1}$, $P^* \leftarrow P_{k+1}$, $C^* \leftarrow C_{k+1}$.

---

## C. Approach for Handling Outside New Data

We mainly discuss our J-RFDL and DJ-RFDL for the inductive representation and classification of outside new data.

**Learning the hybrid salient representations of new data.** After $P^*$ is obtained and given a new sample $x_{new}$, J-RFDL and DJ-RFDL obtain the hybrid salient coefficient vector as

$$\aleph_{new} = P^* x_{new}. \quad (37)$$

**Classification of new data by DJ-RFDL.** Since $P^*$ and $C^*$ are jointly obtained, we can involve the outside new data by them. Note that the classification process of DJ-RFDL is very efficient. For new sample $x_{new}$, its soft label vector $u_{new} \in \mathbb{R}^{c\times 1}$ and hard label $l(x_{new})$ can be obtained as

$$l(x_{new}) = \arg\max_{i\leq c}(u_{new})_i,\ \text{where}$$
$$u_{new} = C^{*T} \aleph_{new},\ \aleph_{new} = P^* x_{new} \quad (38)$$

where $(u_{new})_i$ is the i-th entry of estimated soft vector $u_{new}$, i.e., the largest element in $(u_{new})_i$ decides the label of $x_{new}$.

**Classification of new data by J-RFDL.** Since our J-RFDL fails to learn a classifier jointly, we propose to apply a similar process to obtain a linear classifier $C^*$ based on the computed hybrid salient representation $P^*X$ of data X separately:

$$\langle C, E\rangle = \arg\min_{C,E} \|E\|_{2,1} + \beta\|C\|_F^2,\ s.t.\ H^T = X^T P^{*T} C + E, \quad (39)$$

which can be similarly solved as Eq.(36). After obtaining the linear classifier $C^*$, we can similarly obtain the soft label vector and hard label of each new data $x_{new}$ by Eq. (38).

## V. DISCUSSION: RELATIONSHIP ANALYSIS

We will describe some important relations to our algorithm.

## A. Connection between CF and J-RFDL

Recalling the objective function of J-RFDL in Eq.(8), suppose that $\alpha = 0$ and $\gamma = 0$, Eq.(8) can be transformed into

$$\min_{W,V} \left\| X^T - VW^T X^T \right\|_{2,1} = tr\left( \left( X^T - VW^T X^T \right)^T G \left( X^T - VW^T X^T \right) \right). \quad (40)$$

If we fix the diagonal matrix $G$ to be identity matrix, Eq.(40) becomes the following Frobenius-norm based one:

$$\min_{W,V} tr\left( \left( X^T - VW^T X^T \right)^T \left( X^T - VW^T X^T \right) \right) = \left\| X - XWV^T \right\|_F^2, \quad (41)$$

which is just the problem of CF. In other words, CF is a special example of our J-RFDL. But by replacing the $L_{2,1}$-norm with Frobenius-norm, the metric will lose the robust properties.

## B. Connection between DPL and J-RFDL

We then show the relationship between DPL and J-RFDL. For the objective function of our J-RFDL in Eq.(8), if we constrain $\gamma = 0$, the problem of Eq.(8) can be reduced into

$$\min_{D,P,W,V} \left\| X^T - VW^T X^T \right\|_{2,1} + \alpha \left( \left\| V^T - DPX \right\|_{2,1} + \left\| V \right\|_1 \right), \; s.t. \; W,V \geq 0. \quad (42)$$

Suppose the ideal case that the reconstructed data $VW^T X^T$ is equal to the original data $X^T$ is satisfied, i.e., the reconstruction error $\| X^T - VW^T X^T \|_{2,1}$ is zero, by removing the $L_1$-norm sparse constraint on $V$ and adding the constraint $\|d_i\|_2^2 \leq 1$, we have

$$\min_{D,P,Q} \left\| V^T - DPX \right\|_{2,1} = tr\left( \left( V^T - DPX \right)^T Q \left( V^T - DPX \right) \right), \; s.t. \; \|d_i\|_2^2 \leq 1. \quad (43)$$

If we also fix $Q$ to be the identity matrix, the above problem becomes the Frobenius-norm based one. Note that although our J-RFDL also involves a projection to approximate the codes by embedding similarly as DPL, they are different in three aspects. First, DPL is a supervised method and needs label information of all training samples, so its performance may be restricted due to limited number of labeled samples in practice, while J-RFDL is essentially an unsupervised algorithm. Second, DPL is a structured DL method and directly uses an analysis DL strategy for approximating the coefficients without using any constraint on the coefficients, while J-RFDL can obtain the hybrid salient low-rank and sparse coefficients clearly. Third, DPL performs in the original input space, while J-RFDL conducts the hybrid representation in factorized compressed space and moreover the factorization process can also remove the noise, which can further improve the performance. By using the Frobenius-norm to replace the sparse and robust $L_{2,1}$-norm, DPL will lose the robust property against noise and outliers in given data, and the reconstruction may be inaccurate in reality.

## VI. SIMULATION RESULTS AND ANALYSIS

We conduct the simulations to evaluate the effectiveness of our J-RFDL and DJ-RFDL, along with illustrating the comparison results with the other related representation and classification techniques. We perform all the simulations on a PC with Intel (R) Core (TM) i7-7700 CPU @ 3.6 GHz 8G.

## A. Baseline and Setting

We mainly evaluate J-RFDL for unsupervised representation and evaluate DJ-RFDL for classification. The representation result of our J-RFDL is mainly compared with those of closely related low-rank and/or sparse coding algorithms, i.e., KSVD [20], IRPCA [14], LatLRR [15], rLRR [16] and I-LSPFC [18]. For classification, we compare our J-RFDL and DJ-RFDL with the unsupervised KSVD, IRPCA, LatLRR, I-LSPFC and rLRR, as well as the supervised D-KSVD [25], LC-KSVD1 [27], LC-KSVD2 [27], *Discriminative I-LSPFC* (D-LSPFC) [18], DLSI[21], ADDL [22], DPL [17] and LRSDL [52]. Note that IRPCA, LatLRR, rLRR and I-LSPFC are unsupervised models and cannot classify new data directly, so we employ the same classifier training and classification process by Eqs.(38-39) as our J-RFDL for them for the fair comparison. For KSVD, we similarly compute a classifier as the classifier training process separately as the LC-KSVD1 algorithm.

Four kinds of images, including face images, object images, handwriting digital images and natural scene images, are tested. Details of used databases are shown in Table I. As a common practice, each face image is resized into 32×32 pixels, which corresponds to a 1024-D vector. For classification, we split each dataset into a training set and a test set, where the training set is used for representation and classifier learning, and the test set is used for evaluation. For unsupervised models, the training set has no supervised prior, while all the training samples are labeled for supervised methods. The classification accuracy of each method is finally obtained by comparing the predicted labels with the ground-truth labels provided by data corpus.

**Table I:** Descriptions of Used Image Datasets.

| Dataset Name | # Samples | # Dim | # Classes |
|---|---|---|---|
| COIL20 object [43] | 1440 | 1024 | 20 |
| UMIST face [49] | 1012 | 1024 | 20 |
| CMU PIE face [44] | 11554 | 1024 | 68 |
| MIT CBCL face [57] | 3240 | 1024 | 10 |
| USPS digits [23] | 9298 | 256 | 10 |
| MNIST digits [58] | 70000 | 784 | 10 |
| Optical digits (OHD) [45] | 5620 | 64 | 10 |
| Caltech101 objects [62] | 9144 | 4096 | 101 |
| Fifteen natural scenes [63] | 4485 | 4096 | 15 |

## B. Investigation of Parameters

We analyze the parameter sensitivity of J-RFDL and DJ-RFDL. Since parameter selection is still an open issue, a heuristic way is used to select the most important ones. The COIL20 object database [43] is evaluated. For our J-RFDL, we can explore the effects of parameters $\alpha$ and $\gamma$ on the classification results by grid search. To see the effects of parameters, we randomly select 20 images from each subject for training and test on the rest. We average the results over 10 random splits of training and test samples with varied parameters from the candidate set $\{10^{-9},\ldots,10^1\}$. The results are shown in Fig.2. We find that J-RFDL performs well for a wide range of parameter settings. In this paper, we simply set $\alpha=1$ and $\gamma=10^{-5}$ for J-RFDL.

DJ-RFDL includes three model parameters, so we fix one of them and explore the effects of other two on the result by grid search. To see the effects of different parameters, we first fix $\alpha=10^{-5}$ and tune $\beta$ and $\gamma$ by grid search from the candidate set $\left[10^{-9},10^{-8},\ldots,10^1\right]$ in Fig.3(a), from which we can find that DJ-RFDL with $\beta \geq 10^{-5}$ and $\gamma \leq 10^{-2}$ can deliver good and stable results. We then fix $\gamma=10^{-3}$ to explore the effects of $\beta$ and $\alpha$ in Fig.3(b), from which we can see that DJ-RFDL with $\beta \geq 10^{-5}$ and $\alpha \leq 10^{-1}$ can deliver good results. Finally, we fix $\beta=10^{-3}$ and explore the effects of $\gamma$ and $\alpha$ in Fig.3(c), and we can observe that DJ-RFDL with each $\gamma$ and $\alpha \leq 10^{-1}$ can deliver good and stable results. Note that similar findings can be obtained from other dataset, so we simply select the parameters from the range of $\alpha \leq 10^{-1}$, $\beta \geq 10^{-5}$ and $\gamma \leq 10^{-2}$ for DJ-RFDL in this paper.

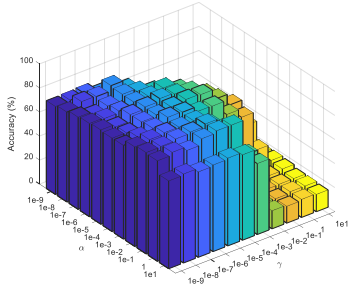

**Fig. 2:** Parameter sensitivity analysis of J-RFDL on COIL20 database.

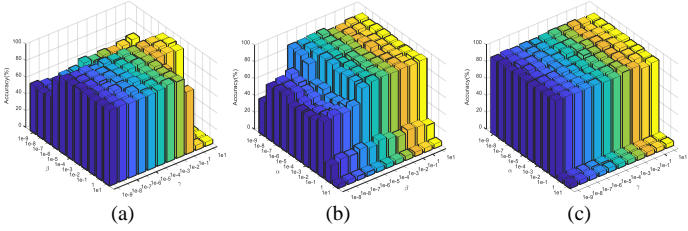

**Fig. 3:** Parameter sensitivity analysis of our DJ-RFDL on COIL20, where (a) fix $\alpha$ to tune $\beta$ and $\gamma$ using grid search; (b) fix $\gamma$ to tune $\beta$ and $\alpha$ using grid search; (c) fix $\beta$ to tune $\gamma$ and $\alpha$ using grid search.

**Table II:** Classification results of J-RFDL under different parameter settings on three databases.

| Parameter settings | MIT CBCL | COIL20 | USPS |
|---|---|---|---|
| $\alpha \neq 0$, $\gamma \neq 0$ | **96.12** | **84.33** | **82.38** |
| $\alpha = 0$, $\gamma \neq 0$ | 8.69 | 4.93 | 11.75 |
| $\alpha \neq 0$, $\gamma = 0$ | 91.96 | 82.88 | 63.19 |

**Table III:** Classification results of DJ-RFDL under different parameter settings on three databases.

| Parameter settings | MIT CBCL | COIL20 | USPS |
|---|---|---|---|
| $\alpha \neq 0$, $\beta \neq 0$, $\gamma \neq 0$ | **98.31** | **94.05** | **82.20** |
| $\alpha = 0$, $\beta \neq 0$, $\gamma \neq 0$ | 40.24 | 33.12 | 52.71 |
| $\alpha \neq 0$, $\beta = 0$, $\gamma \neq 0$ | 91.68 | 66.34 | 57.07 |
| $\alpha \neq 0$, $\beta \neq 0$, $\gamma = 0$ | 92.21 | 90.92 | 19.37 |

In addition to the above parameter analysis, we also explore the effects and contributions of the involved terms associated with the model parameters in the objective functions of J-RFDL and DJ-RFDL by setting the parameters to 0, respectively. In this study, MIT CBCL, COIL20 and USPS databases are used. MIT CBCL database has images of 10 persons, and each person has 324 images. The head models are generated by fitting a morphable model to the high-resolution training images. USPS database has 9298 handwritten digits ('0'-'9') of 16×16 pixels [23]. In this paper, we choose 300 samples from each digit and each digit is represented by a 256-dimensional vector. In this experiment, we select 4, 20 and 40 per class from these datasets respectively for training. The results of J-RFDL and DJ-RFDL are shown in Tables II and III. We can find that when we set $\alpha$, $\beta$ or $\gamma$ to 0, the classification results are all decreased. More specifically, when $\alpha = 0$ in J-RFDL and DJ-RFDL, i.e., the reconstruction term of DL is removed, the result is the lowest. That is, this reconstruction term is more important. When $\gamma = 0$ in J-RFDL and DJ-RFDL, i.e., the hybrid salient representation coefficients are not be obtained any more, the classification result is also decreased. If $\beta = 0$ in DJ-RFDL, i.e., the linear classifier $C$ is not optimized jointly, the classification result is still decreased. This implies that minimizing the classification error based on hybrid salient coefficients at the same time is needed. Hence, the involved several terms in the problems of our methods are all important to improve the performance.

It should be noted that the model parameter(s) of the other compared algorithms are also carefully chosen from the same candidate set as our methods for the fair comparison, and the presented results of other compared algorithms below are also based on the chosen best parameters from candidate set. But due to the page limitation, we will not describe the parameter selection results of other compared algorithms, because this is beyond the main scope of this paper.

### C. Comparison of Actual Computation Time

For clear observation, Fig.4 shows the averaged training and testing time separately on three databases, i.e., UMIST [49], CMU PIE [44] and COIL20 [43]. Each subject of UMIST is described in a range of poses from profile to frontal views. Following the common procedures over CMU PIE, 170 near frontal face images per person are used, containing five near frontal poses (C05, C07, C09, C27, and C29), and all images have different illuminations, lighting, and expressions.

For each database, we randomly select 20 simples from each class for training and test on the rest. We can see from Figs.4(a) and (c) that: (1) DPL and ADDL are slightly faster than other methods, and our J-RFDL and DJ-RFDL are faster than the KSVD, D-KSVD, LC-KSVD1 and LC-KSVD2; (2) J-RFDL, DJ-RFDL, DPL and ADDL spend relatively less time in testing, compared with KSVD, D-KSVD and LC-KSVD. For the result in Fig.4 (b) on a larger face database, KSVD, D-KSVD and LC-KSVD spend much more time in testing than others.

Besides, we also prepare a simulation to evaluate the effects of using the whole dataset instead of selected number of images on the time consumption. UMIST, CMU PIE and COIL20 are also used. That is, all the samples are used for training and we report the training time in Fig.5. For UMIST, the dictionary size is set to 450. For CMU PIE, we use the principal component analysis (PCA) to reduce the dimension by preserving 99% energy. We can find that DPL and ADDL are still faster than other methods, and J-RFDL and DJ-RFDL deliver comparable or even less running time than KSVD, D-KSVD, LC-KSVD1 and LC-KSVD2 algorithms in most cases.

### D. Convergence Analysis Results

We present some convergence results of our proposed J-RFDL and DJ-RFDL. In this study, four image databases, i.e., COIL20, UMIST, MIT CBCL and USPS, are used. For each database, we select 10 images per class for training and the averaged convergence results over 10 splits are shown in Fig.6. We can find that the convergence errors of our J-RFDL and DJ-RFDL are non-increasing and usually converge to a very small value. Moreover, J-RFDL and DJ-RFDL usually converge with the number of iterations ranging from 20 to 60 in most cases.

### E. Face Recognition

We evaluate J-RFDL and DJ-RFDL for face recognition on two public real databases, i.e., CMU PIE [44] and MIT CBCL [57]. Some image examples are illustrated in Fig.7. The number of atoms is set as its maximum for each method.

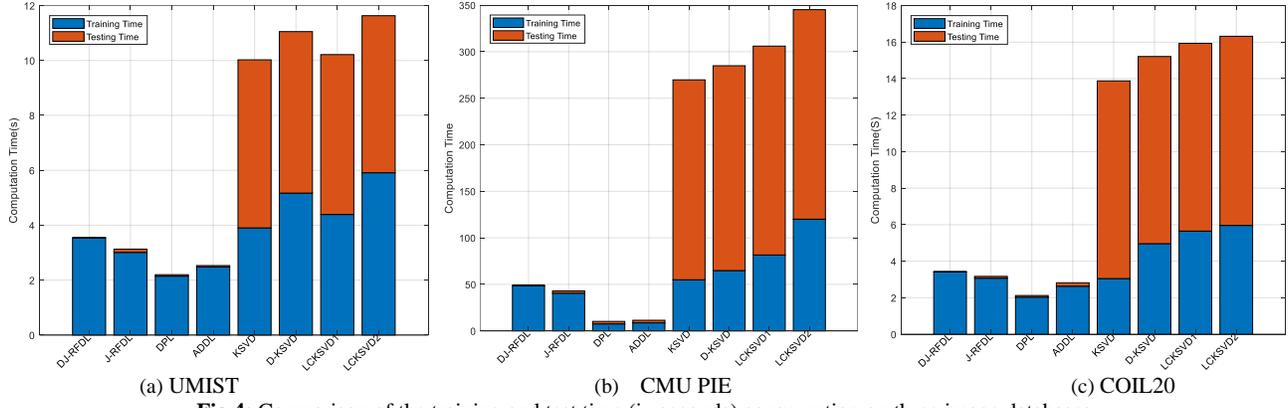
**Fig.4:** Comparison of the training and test time (in seconds) consumption on three image databases.

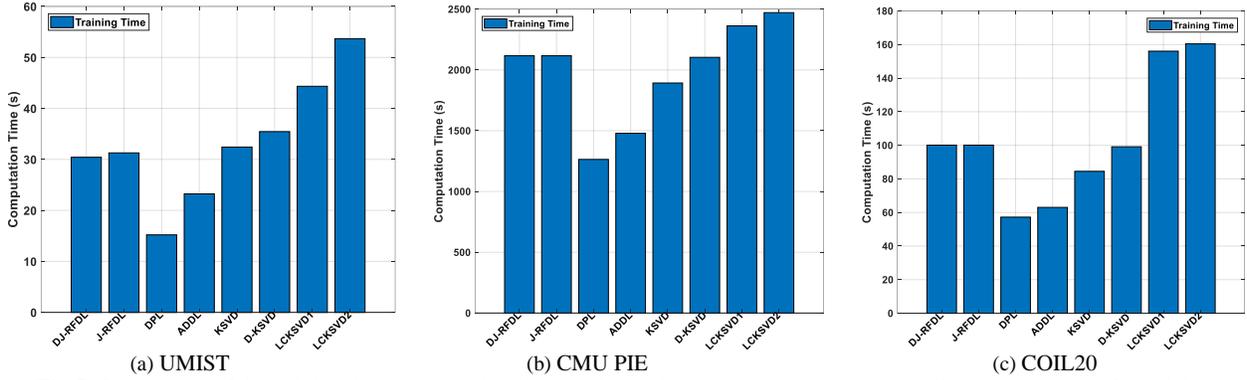
**Fig.5:** Comparison of the training time consumption(in seconds) on three image databases, where all samples are used for training.

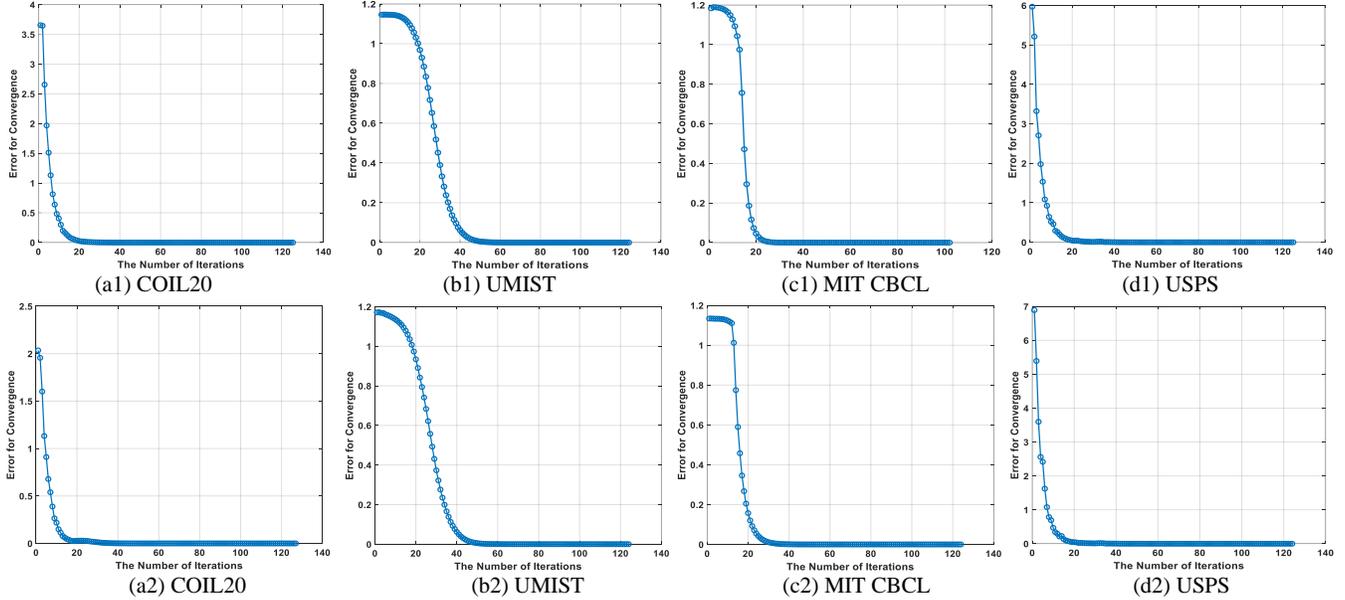
**Fig.6:** The convergence results of our J-RFDL (a1-d1) and DJ-RFDL (a2-d2) on four real-world image databases.

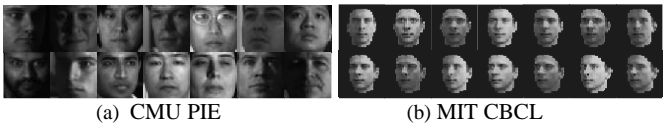

(a) CMU PIE    (b) MIT CBCL

**Fig. 7:** Image examples of the evaluated face databases.

**Recognition results on CMU PIE**. We mainly compare the results of J-RFDL and DJ-RFDL with those of unsupervised KSVD, LatLRR, rLRR, IRPCA, I-LSPFC, and discriminative D-KSVD, DLSI, LC-KSVD, D-LSPFC, DPL and ADDL in this study. Table IV shows the results of each method, where we fix the number (*f*) of training faces per person and average the results over 10 random splits of training/testing images. We report the mean accuracy (%) with the standard deviation (STD) and highest accuracy (%). Note that we show the results by two groups, i.e., unsupervised and supervised ones, and highlight the best record in each group in bold. For unsupervised group, we compare our J-RFDL with KSVD, LatLRR, rLRR, IRPCA, I-LSPFC. $\alpha = 10^{-4}$, $\beta = 10^{-4}$ and $\gamma = 10^{-4}$ are used in DJ-RFDL and we have the following observations. (1) For unsupervised methods, J-RFDL usually obtains better results than compared methods. I-LSPFC and rLRR also work well by obtaining the comparable results to J-RFDL. The superiority of rLRR over LatLRR can be attributed to preserving the manifold structures of features. (2) For supervised group, DJ-RFDL obtains higher accuracies across all numbers of training samples.

In addition, we also compare the average recognition rates by varying the dictionary size $K$. The training set is formed by randomly choosing 30 images per person. $K$ = 340, 680, 1020, 1360, 1700, and 2040 are tested, and the results are illustrated in Fig. 8(a). We see that the rates of each method are increasing along with the increasing numbers of dictionary atoms, and more importantly DJ-RFDL can outperform all its competitors by delivering the enhanced performance. ADDL also performs well by delivering enhanced recognition performance.

Table IV: Performance comparison of the algorithms on the CMU PIE face database.

| Setting Method | CMU PIE (3 train) Mean±STD | Best | CMU PIE (5 train) Mean±STD | Best | CMU PIE (7 train) Mean±STD | Best | CMU PIE (9 train) Mean±STD | Best |
|---|---|---|---|---|---|---|---|---|
| IRPCA | 49.15±1.82 | 52.51 | 58.61±1.69 | 61.29 | 64.59±1.35 | 66.55 | 73.48±1.02 | 75.89 |
| LatLRR | 53.95±1.08 | 56.21 | 60.41±0.93 | 61.26 | 70.69±0.99 | 72.04 | 76.42±1.29 | 78.95 |
| rLRR | 54.13±2.21 | **56.96** | 64.36±1.32 | 66.42 | 74.38±0.63 | 75.04 | 78.56±0.89 | **80.52** |
| KSVD | 48.22±1.73 | 52.04 | 55.31±1.27 | 57.33 | 66.34±0.62 | 68.21 | 75.59±0.33 | 77.12 |
| I-LSPFC | 51.82±0.38 | 52.10 | 63.21±1.17 | 65.15 | 70.11±1.15 | 72.28 | 74.54± 0.70 | 75.26 |
| J-RFDL | **55.41±0.65** | 56.15 | **68.55±1.91** | **71.19** | **74.71± 0.84** | **75.68** | **79.06±0.73** | 79.78 |
| DLSI | 33.10±1.47 | 34.99 | 49.13±1.19 | 50.71 | 64.13±1.03 | 65.77 | 74.09±1.16 | 75.67 |
| ADDL | 57.59±1.42 | 59.11 | 68.84±1.35 | 70.15 | 76.49±0.56 | 77.33 | 80.82±0.66 | 81.55 |
| LC-KSVD1 | 52.38±1.35 | 54.04 | 63.75±1.43 | 66.27 | 70.33±0.52 | 70.09 | 76.92±0.72 | 77.66 |
| LC-KSVD2 | 55.12±1.24 | 56.98 | 67.94±1.31 | 68.26 | 73.33±0.72 | 73.83 | 77.62±0.48 | 78.53 |
| DPL | 57.04±1.65 | 59.03 | 70.34±1.39 | 71.98 | 77.13±2.33 | 79.16 | 82.41± 0.67 | 83.69 |
| D-KSVD | 50.17±3.69 | 54.22 | 59.35±3.92 | 64.54 | 67.82±0.89 | 69.36 | 76.51±1.10 | 77.76 |
| D-LSPFC | 58.05±1.25 | 60.48 | 68.39±0.65 | 69.14 | 76.89± 0.81 | 78.36 | 81.21±0.19 | 82.07 |
| LRSDL | 55.86±2.37 | 58.42 | 67.18±1.21 | 70.35 | 77.22±0.91 | 89.27 | 79.60±1.23 | 82.47 |
| DJ-RFDL | **61.08±1.41** | **63.73** | **73.08±0.86** | **74.29** | **80.75± 0.58** | **81.81** | **84.40±0.91** | **85.72** |

Table V: Performance comparison of the algorithms on the MIT CBCL face database.

| Setting Method | MIT CBCL (1 train) Mean±STD | Best | MIT CBCL (2 train) Mean±STD | Best | MIT CBCL (3 train) Mean±STD | Best | MIT CBCL (4 train) Mean±STD | Best |
|---|---|---|---|---|---|---|---|---|
| IRPCA | 62.00±3.50 | **67.46** | 82.18±4.79 | 88.59 | 92.35±3.13 | 96.79 | 95.25±3.75 | 99.28 |
| LatLRR | 51.07±2.09 | 55.29 | 63.99±4.70 | 68.94 | 77.16±3.42 | 81.45 | 80.32±3.30 | 85.52 |
| rLRR | 60.71±3.57 | 65.29 | 70.78±6.30 | 84.65 | 85.29±4.16 | 90.69 | 84.51±3.07 | 88.72 |
| KSVD | 55.41±1.23 | 57.36 | 62.57±2.31 | 64.39 | 86.89±6.17 | 91.90 | 93.15±2.42 | 95.90 |
| I-LSPFC | 57.71±3.68 | 66.59 | 83.70±4.08 | **89.53** | 92.50±3.60 | 96.70 | 95.95±1.41 | 97.52 |
| J-RFDL | **63.81±3.27** | 66.84 | **85.91±2.33** | 87.33 | **93.65±3.13** | **97.07** | **96.12±2.30** | **99.59** |
| DLSI | 60.44±3.14 | 65.32 | 74.72±3.29 | 77.08 | 89.31±3.37 | 93.42 | 89.50±4.24 | 93.49 |
| ADDL | 62.50±2.62 | 65.83 | 84.42±5.71 | 92.12 | 93.23±2.74 | 96.91 | 97.25±1.62 | 99.42 |
| LC-KSVD1 | 59.27±3.67 | 63.23 | 80.17±4.33 | 85.76 | 90.13±3.47 | 95.52 | 93.33±3.34 | 95.43 |
| LC-KSVD2 | 61.83±2.45 | 64.31 | 83.35±5.63 | 89.88 | 92.11±3.67 | 96.32 | 94.15±3.05 | 96.37 |
| DPL | 60.81±1.81 | 62.47 | 85.25±4.73 | 91.42 | 93.81±1.37 | 96.14 | 96.53±1.62 | 99.50 |
| D-KSVD | 58.23±3.23 | 63.21 | 84.15±6.16 | 91.43 | 91.69±3.78 | 94.17 | 95.62±1.61 | 97.19 |
| D-LSPFC | 62.03±2.81 | 67.33 | 85.66±4.26 | 92.31 | 93.91±3.38 | **98.01** | 96.47±1.48 | 98.96 |
| LRSDL | 59.05±2.42 | 64.19 | 83.61±3.96 | 89.57 | 93.42±2.82 | 97.68 | 97.41±1.28 | 98.49 |
| DJ-RFDL | **64.36±3.19** | **70.77** | **86.83±5.32** | **95.59** | **96.01±1.51** | 97.76 | **98.31±1.45** | **100** |

**Recognition results on MIT CBCL.** We test each method for recognizing the faces of MIT CBCL. Table V describes the result of each method, where we fix the number ($f$) of training faces of each person (i.e., $f$=1, 2, 3, 4) and average the results over 10 random splits. We show the mean and highest accuracy (%) for each method, and also divide the results into two groups according to the unsupervised and supervised methods. We find that J-RFDL and DJ-RFDL deliver comparable or better results than the other algorithms in each group. $\alpha = 10^{-4}$, $\beta = 10^{-4}$ and $\gamma = 10^{-4}$ are applied for our DJ-RFDL for simulations.

Additionally, we randomly choose 6 samples from each class as training set and evaluate each DL method with varying size $K$ of dictionary, i.e., $K$ = 20, 30, 40, 50, 60, in Fig. 8(b) that indicates that our DJ-RFDL maintains a higher recognition rate than other methods even when the dictionary size is relatively small. ADDL also performs well by delivering better results.

### F. Handwriting Recognition

We also evaluate each method for handwriting recognition on three public handwritten databases, i.e., USPS [23] and OHD [45] and MNIST, and some image examples are shown in Fig.9.

**Recognition results on USPS.** We first evaluate each model to recognize the handwritten digits of USPS. Table VI shows the recognition result of each approach with varying numbers

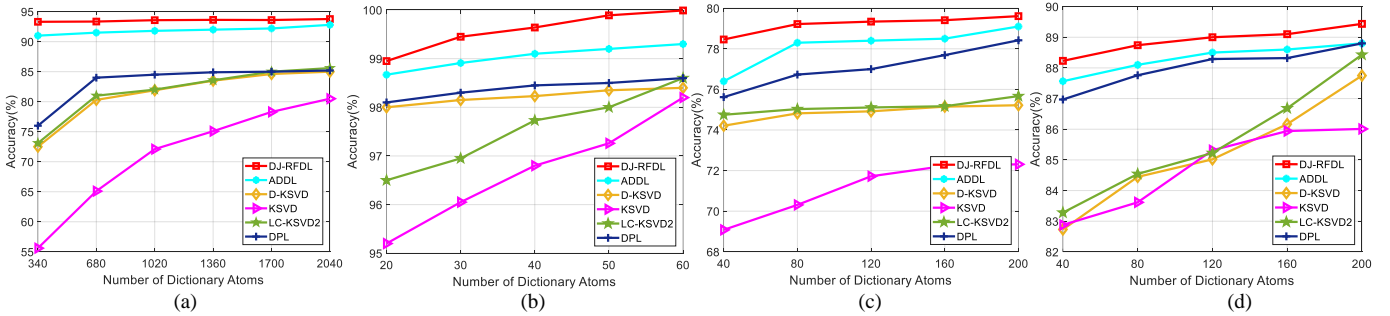
**Fig. 8:** Classificaton performance with varying dictionary sizes on (a) CMU PIE, (b) MIT CBCL, (c) USPS and (d) OHD databases.

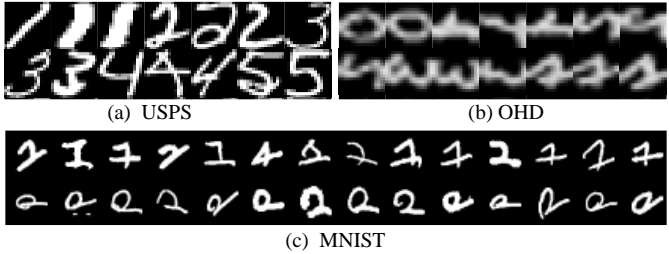
**Fig. 9:** Image examples of the evaluated handwritten databases.

of training data, i.e., 10, 20, 30 and 40 images per class, and test on the rest. The averaged results are shown in Table VI, where we report the mean accuracy with STD and highest record, and also show the results using two groups. $\alpha = 10^{-1}$, $\beta = 10^{-3}$ and $\gamma = 10^{-6}$ are used for our DJ-RFDL. We can find that: (1) for unsupervised methods, J-RFDL obtains the highest accuracy on average, especially when the number of training data is small. (2) For discriminative group, DJ-RFDL outperforms the other algorithms across all the training numbers, especially when the number of training samples is small. ADDL also works well.

To evaluate DJ-RFDL with varying dictionary sizes $K$=40, 80, 120, 160, 200. We randomly select 20 images per class to form the training set and test on the rest. The results under varying dictionary sizes are shown in Fig. 8(c). We can see that our methods can obtain higher accuracies than other methods.

**Recognition results on OHD database.** This database has handwritten digits ('0'-'9') of 8×8 pixels where each element is an integer in the range from 0 to 16. Table VII shows the result of each method with varying numbers of training data, i.e., 3, 6, 9 and 12 images per class. The results are averaged over 10 random splits of training and test digits. $\alpha = 10^{-5}$, $\beta = 10^{-5}$ and $\gamma = 10^{-7}$ are used for DJ-RFDL. We find that for unsupervised and discriminative group, our methods can obtain better results than other methods, and the difference between ours and other models is larger when the number of training samples is small.

**TABLE VI:** Performance comparison of each algorithm on the USPS database.

| Setting<br>Method | USPS (10 train) | | USPS (20 train) | | USPS (30 train) | | USPS (40 train) | |
|---|---|---|---|---|---|---|---|---|
| | Mean±STD | Best | Mean±STD | Best | Mean±STD | Best | Mean±STD | Best |
| IRPCA | 62.52±2.16 | 65.17 | 70.47±2.48 | 73.96 | 72.88±1.82 | 76.81 | 76.75±1.61 | 79.73 |
| LatLRR | 63.18±2.18 | 66.17 | 71.04±1.87 | 73.08 | 73.31±0.84 | 74.52 | 75.22±0.94 | 76.58 |
| rLRR | 65.80±2.09 | **69.52** | 72.17±2.31 | **75.11** | 74.23±2.06 | 77.11 | 76.83±0.91 | 79.73 |
| KSVD | 64.62±1.47 | 67.14 | 72.21±1.69 | 72.94 | 77.55±1.59 | 80.22 | 80.25±0.71 | 81.62 |
| I-LSPFC | 63.87±1.97 | 66.97 | 71.39±2.28 | 74.29 | 76.00±1.57 | 77.85 | 78.37±0.91 | 80.19 |
| J-RFDL | **66.00±2.05** | 68.83 | **74.06±0.84** | 74.89 | **78.38±1.13** | **80.26** | **82.38±0.36** | **83.08** |
| DLSI | 67.47±1.78 | 70.12 | 74.69±1.29 | 77.61 | 78.59±1.27 | 80.91 | 82.42±1.12 | 83.47 |
| ADDL | 69.71±1.72 | 72.52 | 78.35±1.59 | 80.21 | 80.79±0.63 | 81.82 | 83.06±1.14 | 84.92 |
| LC-KSVD1 | 63.07±0.41 | 64.09 | 74.88±0.73 | 76.00 | 77.83±0.61 | 78.52 | 78.83±2.16 | 80.96 |
| LC-KSVD2 | 67.90±1.28 | 68.75 | 75.28±1.99 | 77.57 | 79.42±1.17 | 80.04 | 81.79±0.56 | 82.69 |
| DPL | 68.75±1.93 | 71.45 | 77.43±1.06 | 78.68 | 80.37±0.59 | 81.26 | 82.75±0.72 | 83.77 |
| D-KSVD | 65.58±1.82 | 68.10 | 75.05±1.71 | 76.46 | 78.87±0.58 | 79.37 | 80.91±0.68 | 81.65 |
| D-LSPFC | 66.58±1.63 | 69.24 | 74.17±2.21 | 76.89 | 78.69±2.94 | 82.15 | 81.74±2.41 | 84.19 |
| LRSDL | 69.14±1.67 | 72.58 | 77.53±2.84 | 80.57 | 81.81±0.57 | 82.94 | 83.08±1.24 | 85.17 |
| DJ-RFDL | **73.12±0.87** | **74.52** | **79.35±2.13** | **81.61** | **82.94±0.86** | **84.00** | **84.20±1.05** | **85.58** |

In addition, to evaluate our DJ-RFDL with varying size $K$ of dictionary, we randomly select 20 digit images per class as the training set and select the dictionary size as $K$=40, 80, 120, 160, 200. The results are shown in Fig. 8(d). We see that the average rate of each method is increased with the increasing numbers of atoms. DJ-RFDL can outperform its other competitors in most cases, and works steady with varying dictionary sizes.

**Recognition results on MNIST database.** We also examine each method for recognizing the handwritten digits of MNIST. The MNIST handwritten database has a training set of 60000 samples and a test set of 10000 samples, where each digital image has 28×28 pixels, which can be represented by using a 784-dimensional vector. In this present work, the training set is employed and we extract random features with the dimension being 100 when evaluating the performance of each method. In our simulations, the number of training samples per digit class is set to 60 (totally 600 training samples) and test on the rest. The classification results are described in Table VIII. We can observe that our proposed formulation can deliver the enhanced results compared with the other related methods.

TABLE VII: Performance comparison of each algorithm on the OHD database.

| Method | OHD (3 train) Mean±STD | Best | OHD (6 train) Mean±STD | Best | OHD (9 train) Mean±STD | Best | OHD (12 train) Mean±STD | Best |
|---|---|---|---|---|---|---|---|---|
| IRPCA | 64.13±3.97 | 70.57 | 69.67±3.53 | 75.85 | 78.23±2.54 | 81.65 | 83.03±1.73 | **86.04** |
| LatLRR | 64.06±4.50 | 71.32 | 76.96±2.12 | 80.75 | 79.61±1.70 | 82.14 | 81.53±2.13 | 84.00 |
| rLRR | 68.41±2.74 | 72.99 | 78.84±2.15 | **82.48** | 81.36±1.93 | 83.62 | 84.97±0.76 | 85.67 |
| KSVD | 61.67±4.56 | 67.47 | 67.21±3.28 | 70.84 | 72.17±3.22 | 75.73 | 76.89±2.94 | 80.31 |
| I-LSPFC | 72.46±4.26 | **79.75** | 78.59±3.77 | 81.52 | 80.29±3.62 | **85.13** | 83.12±1.64 | 84.73 |
| J-RFDL | **73.40±0.47** | 74.22 | **79.24±0.42** | 80.01 | **82.54±0.33** | 83.15 | **85.22±0.21** | 85.82 |
| DLSI | 60.43±4.41 | 65.72 | 68.43±3.26 | 71.24 | 73.32±3.58 | 77.51 | 78.56±3.24 | 82.37 |
| ADDL | 75.26±2.34 | **79.33** | 79.67±1.89 | 81.80 | 82.87±1.42 | 84.46 | 85.51±0.35 | 86.21 |
| LC-KSVD1 | 62.71±2.34 | 65.02 | 68.75±4.21 | 73.06 | 72.69±4.68 | 78.34 | 77.36±3.27 | 82.23 |
| LC-KSVD2 | 63.69±2.11 | 66.47 | 69.95±3.95 | 74.21 | 75.55±3.68 | 80.43 | 78.12±3.62 | 82.31 |
| DPL | 72.24±4.34 | 77.36 | 79.62±2.89 | 82.39 | 81.66±1.84 | 83.01 | 84.74±1.62 | 86.16 |
| D-KSVD | 62.57±3.52 | 67.23 | 68.26±3.21 | 71.26 | 73.74±2.82 | 76.52 | 78.91±2.11 | 80.83 |
| D-LSPFC | 74.28±3.76 | 79.00 | 80.64±1.94 | **83.92** | 81.18±2.45 | **85.18** | 85.43±0.94 | 86.60 |
| LRSDL | 68.94±4.37 | 72.46 | 75.81±2.64 | 79.18 | 79.63±2.57 | 82.11 | 81.28±1.89 | 84.20 |
| DJ-RFDL | **75.91±0.25** | 76.45 | **80.75±0.18** | 80.99 | **83.25±0.16** | 83.51 | **87.60±0.14** | **87.82** |

TABLE VIII: Performance comparison of each algorithm on the MNIST handwriting digit database.

| Evaluated Methods | Mean±STD |
|---|---|
| IRPCA (600 training samples) | 65.55±2.17 |
| LatLRR (600 training samples) | 64.84±1.42 |
| rLRR (600 training samples) | 68.62±1.32 |
| KSVD (600 training samples) | 66.58±1.42 |
| I-LSPFC (600 training samples) | 68.24±2.11 |
| **J-RFDL** (600 training samples) | **70.93**±1.97 |
| DLSI (600 training samples) | 70.22±1.88 |
| ADDL (600 training samples) | 74.33±2.21 |
| LC-KSVD1 (600 training samples) | 70.11±2.58 |
| LC-KSVD2 (600 training samples) | 72.55±2.37 |
| DPL (600 training samples) | 73.81±1.47 |
| D-KSVD (600 training samples) | 72.41±2.32 |
| D-LSPFC (600 training samples) | 73.24±1.79 |
| LRSDL (600 training samples) | 73.41±2.49 |
| **DJ-RFDL** (600 training samples) | **76.76**±1.36 |

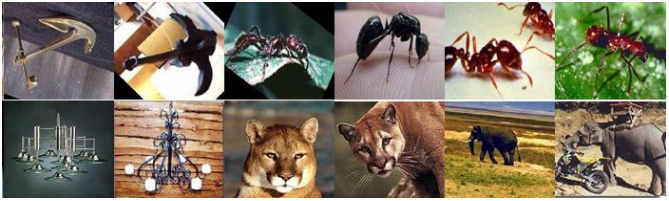

**Fig. 10:** Image examples of the Caltech101 database.

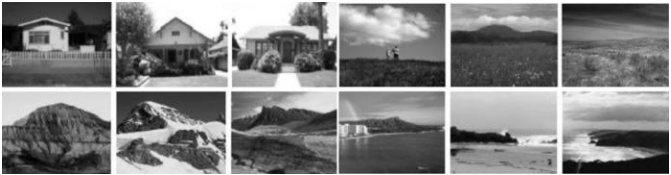

**Fig. 11:** Image examples of the fifteen natural scene categories database.

*G. Natural Object Image Recognition*

In this simulation, we evaluate each algorithm for recognizing the object images of the popular Caltech101 database [62]. This database has 9144 images, consisting of 101 object classes and 1 background class. The number of images in each class varies from 31 to 800 and the size of each image is roughly 300×200 pixels. Some image examples are show in Fig.10. In this study, we evaluate each method using deep convolutional features, and we employ the classical Alexnet framework [64] to extract the deep features from original images. To extract deep features, the input size of images is set according to the requirement Alexnet [64] and the features of the penultimate full connection layer ('fc7') are evaluated. Finally, the dimension of extracted features is 4096. Then, we can construct a feature matrix based on extracted deep features for representation learning by each method for fair comparison. The number of training samples is set to 10 per class and we test on the rest. The dictionary size is set to the number of training samples. The averaged recognition results are described in Table IX. We see from the results that our methods deliver higher recognition results.

TABLE IX: Performance comparison of each algorithm on the Caltech101 database.

| Evaluated Methods | Mean±STD |
|---|---|
| IRPCA (10 training samples per class) | 70.53±1.89 |
| LatLRR (10 training samples per class) | 71.48±2.11 |
| rLRR (10 training samples per class) | 74.89±1.58 |
| KSVD (10 training samples per class) | 75.11±1.55 |
| I-LSPFC (10 training samples per class) | 74.71±1.72 |
| **J-RFDL**(10 training samples per class) | **76.00±1.51** |
| DLSI (10 training samples per class) | 75.43±2.81 |
| ADDL (10 training samples per class) | 84.23±1.57 |
| LC-KSVD1 (10 training samples per class) | 76.30±1.48 |
| LC-KSVD2(10 training samples per class) | 77.94±1.67 |
| DPL (10 training samples per class) | 83.45±1.49 |
| D-KSVD (10 training samples per class) | 76.71±1.53 |
| D-LSPFC (10 training samples per class) | 82.91±1.35 |
| LRSDL (10 training samples per class) | 83.88±1.59 |
| **DJ-RFDL** (10 training samples per class) | **85.05±1.42** |

TABLE X: Performance comparison of each methods on the fifteen natural scene categories database.

| Evaluated Methods | Mean±STD |
|---|---|
| IRPCA (20 training samples per class) | 49.87±1.76 |
| LatLRR (20 training samples per class) | 50.29±1.88 |
| rLRR (20 training samples per class) | 51.23±1.77 |
| KSVD (20 training samples per class) | 52.63±1.28 |
| I-LSPFC (20 training samples per class) | 52.23±1.45 |
| **J-RFDL**(20 training samples per class) | **54.79±1.31** |
| DLSI (20 training samples per class) | 56.11±1.33 |
| ADDL (20 training samples per class) | 63.18±1.10 |
| LC-KSVD1 (20 training samples per class) | 57.21±1.37 |
| LC-KSVD2(20 training samples per class) | 59.39±2.11 |
| DPL (20 training samples per class) | 62.78±1.27 |
| D-KSVD (20 training samples per class) | 58.19±1.66 |
| D-LSPFC (20 training samples per class) | 61.27±2.14 |
| LRSDL (20 training samples per class) | 62.38±1.77 |
| **DJ-RFDL** (20 training samples per class) | **63.68±1.58** |

*H. Natural Scene Category Recognition*

We also evaluate each method for recognizing the natural scene categories. The fifteen natural scene categories database [63] is involved in this study, which contains 15 natural scenes, that is, suburb, open country, mountain, coast, highway, forest, store, office, kitchen, industrial, living room, bedroom, tall building, street and inside city. Each category has 200 to 400 images that are about 250×300 resolution. Some image examples are show in Fig.11. We also use the deep features exacted by Alexnet [64] for recognition, and select 20 samples per category as training set. Since the inputs of Alexnet are RGB images, in this study we convert each gray image into a RGB image by copying the gray image into each of the three channels of RGB space. The recognition results are shown in Table X, where we find that our models deliver higher accuracies in the two groups.

*I. Noisy Image Recognition Evaluation*

We evaluate the robustness of each method using noisy images. The results under different pixel corruptions are computed. The default dictionary size is set to the number of training samples. Three kinds of real images are evaluated.

*(1) Face recognition with pixel corruptions.* UMIST face database [49] is evaluated. We fix the number of training face images of each class to 6 and vary the percentages of corrupted pixels. Fig.12 shows the result of each method as a function of random corruptions, where the results are averaged based on 10

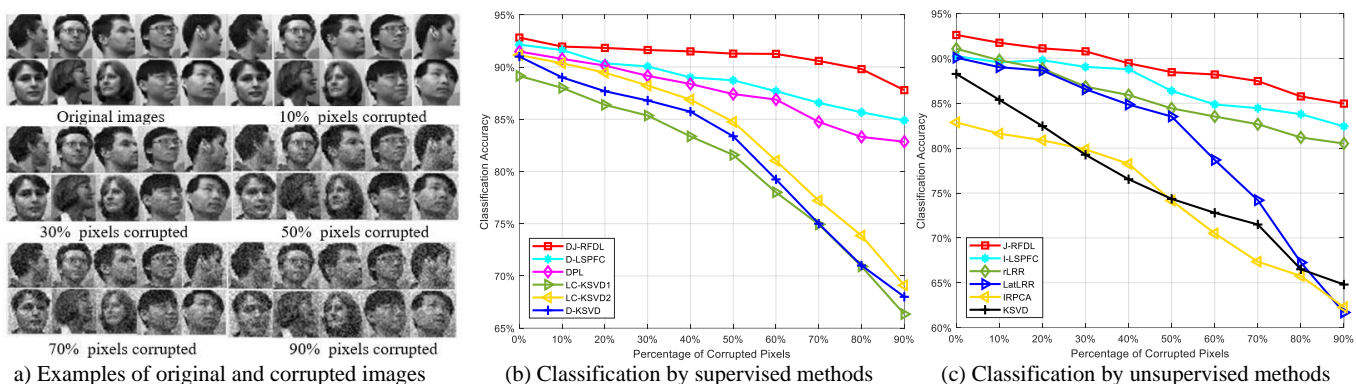

(a) Examples of original and corrupted images  (b) Classification by supervised methods  (c) Classification by unsupervised methods

**Fig. 12:** Handwritten digit recognition results of each algorithm under different levels of pixel corruptions on the UMIST face database.

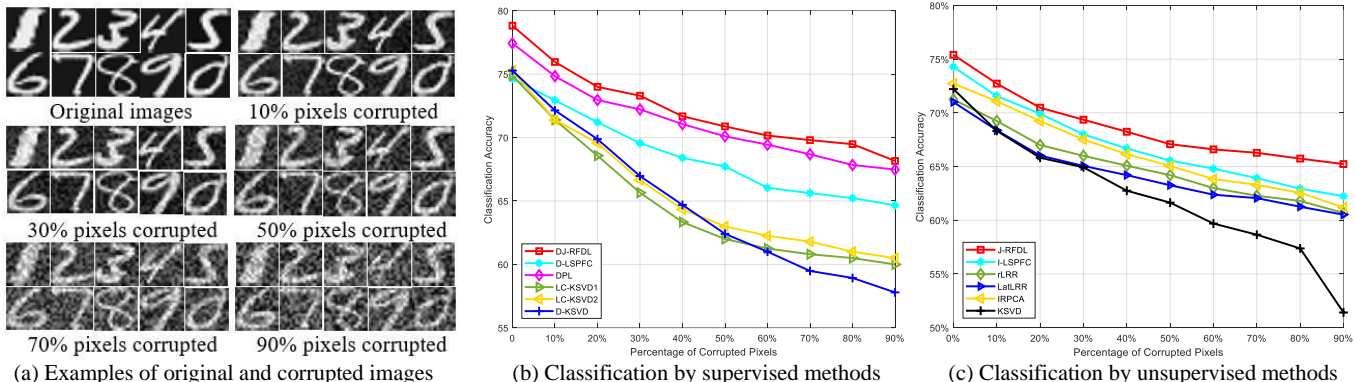

(a) Examples of original and corrupted images  (b) Classification by supervised methods  (c) Classification by unsupervised methods

**Fig. 13:** Handwritten digit recognition results of each algorithm under different levels of pixel corruptions on the USPS digits database.

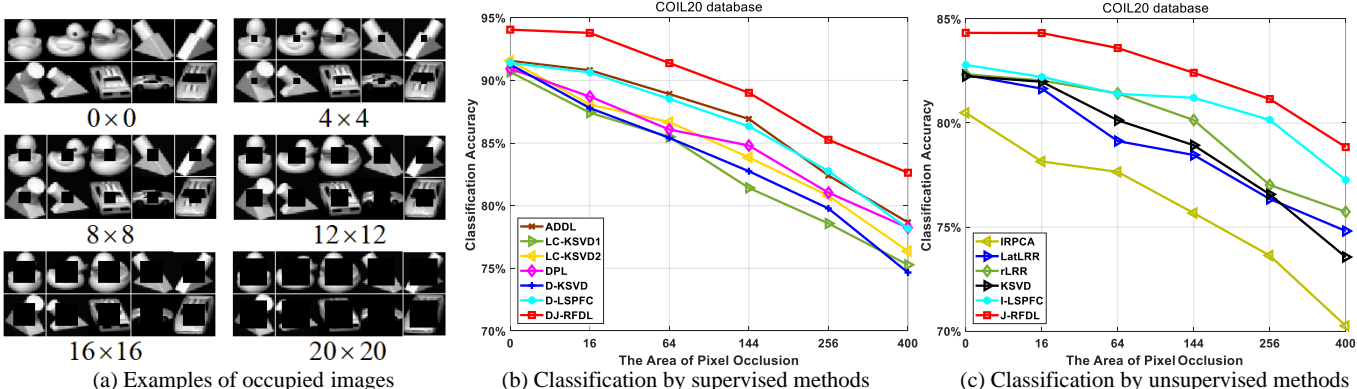

(a) Examples of occupied images  (b) Classification by supervised methods  (c) Classification by unsupervised methods

**Fig. 14:** Object recognition results of each algorithm under different areas of block occlusion on the COIL20 object database.

random splits of training/testing images and 10 random pixel selections. We can see that the performance of each method is decreased with the increasing percentages of corruptions, and the results of our methods usually go down slower than other methods as the percentage of corruptions is increased. For the supervised group, DPL and D-LSPFC are more robust to the noise than other methods. The results of D-KSVD, LC-KSVD1 and LC-KSVD2 are decreased rapidly when the percentage of the corrupted pixels is larger than 50%. For unsupervised group, J-RFDL achieves higher and more stable results than the other compared methods. rLRR and I-LSPFC also perform well.

*(2) Handwritten recognition with pixel corruptions.* We also examine J-RFDL and DJ-RFDL for classifying the handwritten digits under noisy case on USPS database [23]. We select 20 images per digit as training set and test on the rest, and vary the percentage of corrupted pixels from 10% to 90%. Fig.13 shows the averaged results over 10 random splits, from which we find that our J-RFDL and DJ-RFDL outperform other competitors in each group under different percentages of corrupted pixels. For the supervised group, DPL also works well, followed by D-LSPFC, and both are superior to other remaining methods. For the unsupervised group, I-LSPFC and IRPCA also deliver better results, and rLRR also obtains better results than LatLRR. K-SVD is still the worst method in most cases.

*(3) Object recognition with missing pixel values.* We test each approach for classifying the object images under noisy case on the COIL-20 database. We choose 20 images per object as training set and test on the rest. In this study, each method performs classification over the images with different areas of missing pixel values by block occlusion similarly as [51]. More specifically, we define areas with different sizes and then set the pixel values of these areas to 0, i.e., missing values. Fig.14 shows the results of each method as a function of the areas of block occlusion, where some examples of the occupied images are also shown. From the results, we see that the performance of each algorithm is still decreased with the increasing numbers of missing values. Our J-RFDL and DJ-RFDL can deliver higher results than other compared methods in each group.

*J. Further Evaluation on Classification*

We also conduct a simulation to compare our algorithms with several deep networks-based methods, including *Deep Belief Network* (DBN) [53], *Discriminative deep DBN* (DDBN) [54], *Deep Dictionary Learning* (DDL) [55] and *Supervised Deep Dictionary Learning* (SDDL) [56]. Following the procedures in [56], YaleB face database [61] and AR face database [60] are employed as examples. More specifically, we select one half of the images for training and the remaining ones for testing on YaleB. For AR, we randomly chose 20 samples per class for training and test on the rest. The averaged classification results are shown in Table XI, where the results of DBN, DDBN, DDL and SDDL are directly adopted from [56]. From the results, we can find that our proposed DJ-RFDL still outperforms the deep networks-based methods for face recognition. Recent SDDL method obtains comparable results to our method.

VII. CONCLUSION AND FUTURE WORK

We have discussed the problem of recovering the hybrid salient low-rank and sparse representation through robust dictionary learning in a factorized compressed feature space, and present a robust factorization and projective dictionary learning model. J-RFDL clearly unifies the robust matrix factorization, robust dictionary learning and the hybrid salient representation. More specifically, our method aims at improving the representation ability by enhancing the robustness of the reconstruction metric to outliers and noise, encoding the reconstruction errors more accurately and obtaining the hybrid salient coefficients that can best represent given data. By combining the classification error over hybrid salient coefficients into J-RFDL to form a unified model, we also present a discriminative J-RFDL model.

We have examined the effectiveness of our methods based on several public databases. The recognition results based on the original and corrupted samples show the superior performance of our methods, compared with several related approaches. In future, we will extend our methods to semi-supervised scenario by using labeled and unlabeled data. Two possible approaches can be considered. (1) We can compute the salient features and hybrid salient coefficients by minimizing the reconstruction error based on labeled samples within each class, and as the same time minimize the reconstruction error over all classes as a whole to preserve the global structures; (2) We can re-define the initial label matrix $H = [h_1, h_2, \ldots h_N] \in \mathbb{R}^{c \times N}$ based on labeled and unlabeled training samples (i.e., for each training sample $x_j$, $h_{i,j} = 1$ if $x_j$ belongs to the class $i$, $1 \le i \le c$, and else $h_{i,j} = 0$), and at the same time consider propagating the label information from labeled data to the unlabeled data. In addition, the optimal determination of dictionary size still remains an open problem and will also be investigated in future work.

ACKNOWLEDGMENT

The authors would like to express sincere thanks to reviewers for their insightful comments, making our manuscript a higher standard. This work is partially supported by National Natural Science Foundation of China (61672365, 61732008, 61725203, 61622305, 61871444 and 61572339), and the Fundamental Research Funds for the Central Universities of China (JZ2019-HGPA0102). Dr. Zhao Zhang is the corresponding author.

**TABLE XI:** Comparison Results on YaleB and AR.

| Methods | YaleB face | AR face |
| --- | --- | --- |
| DBN | 34.91±4.14 | 43.62±4.32 |
| DDBN | 38.20±4.57 | 60.34±5.68 |
| DDL | 92.66±5.11 | 93.35±5.26 |
| SDDL (1-layer) | 89.45±5.52 | 91.13±5.61 |
| SDDL (2-layer) | 90.31±5.27 | 91.17±5.28 |
| SDDL (3-layer) | 96.50±5.36 | 94.57±5.41 |
| SDDL (4-layer) | 96.16±5.57 | 94.24±5.62 |
| **DJ-RFDL** | **97.50±3.31** | **98.33+3.27** |

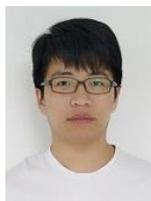
**Jihuan Ren** is currently working toward the research degree at School of Computer Science and Technology, Soochow University, Suzhou 215006, China. His current research interests include pattern recognition, machine learning, data mining and their applications. Specifically, he is very interested in designing the advanced low-rank coding and dictionary learning algorithms for the robust image representation and classification.

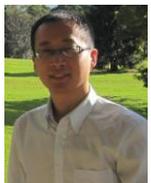
**Zhao Zhang** (SM'17- ) is a Full Professor in the School of Computer Science, Hefei University of Technology, Hefei, China. He received the Ph.D. degree from the Department of Electronic Engineering at City University of Hong Kong, in 2013. His current research interests include Data Mining & Machine Learning, Image Processing & Pattern Recognition. He has authored/co-authored more than 90 technical papers published at prestigious journals and conferences, such as IEEE TIP (6), IEEE TKDE (6), IEEE TNNLS (8), IEEE TSP, IEEE TCSVT, IEEE TCYB, IEEE TBD, IEEE TII (2), ACM TIST, Pattern Recognition (6), Neural Networks (8), Computer Vision and Image Understanding, ACM Multimedia, IJCAI, ICDM (4), SDM, ICASSP and ICMR, etc. Specifically, he has published 26 regular papers in IEEE/ACM Transactions. Dr. Zhang is serving/served as an Associate Editor (AE) for IEEE Access, Neurocomputing and IET Image Processing. Besides, he has been acting as a Senior PC member/Area Chair of IJCAI、ECAI、BMVC and PAKDD, and a PC member for 10+ popular prestigious conferences (e.g.,CVPR、ICCV、IJCAI、AAAI、ACM MM、ECCV、ICDM、CIKM and SDM). He is now a Senior Member of the IEEE, and a Senior Member of the CCF.

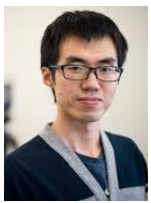
**Sheng Li** (M'17-SM'19-) received the Ph.D. degree from Northeastern University, Boston, MA, in 2017. He is now a Tenure-Track Assistant Professor in the Department of Computer Science at University of Georgia. He has published over 50 papers at leading conferences and journals. He has received the best paper awards (or nominations) at SDM 2014, IEEE ICME 2014, and IEEE FG 2013. He is serving on the Editorial Boards of Neural Computing and Applications, and also serves as an Associate Editor of IEEE Computational Intelligence Magazine, Neurocomputing and IET Image Processing, etc. He has also served as a reviewer for several IEEE Transactions, and program committee member for NIPS, IJCAI, AAAI, and KDD. His research interests include robust machine learning, dictionary learning, visual intelligence, and behavior modeling. He is now a Senior Member of the IEEE.

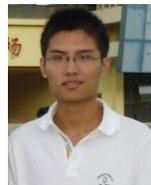
**Yang Wang** is now a Professor at the Hefei University of Technology, China. He earned the PhD degree from the University of New South Wales, Australia. He has published over 50 research papers in pattern recognition and machine learning field, such as IEEE TIP, IEEE TNNLS, IEEE TCSVT, IEEE TMM, IEEE TKDE, IEEE TCYB, IJCAI, ACM SIGIR, ACM Multimedia, IEEE ICDM, ACM CIKM, VLDB Journal, Pattern Recognition and Neural Networks. Yang Wang is an Associate Editor for ACM Transactions on Information Systems (ACM TOIS), while served as a Guest Editor on IEEE Multimedia magazine, Pattern Recognition Letters.

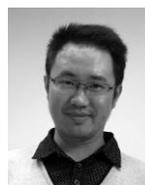
**Guangcan Liu** (M'11-SM'18- ) received the bachelor's degree in mathematics and the Ph.D. degree in computer science and engineering from the Shanghai Jiao Tong University, Shanghai, China, in 2004 and 2010, respectively. He was a Post-Doctoral Researcher with the National University of Singapore, Singapore, from 2011 to 2012, the University of Illinois at Urbana-Champaign, Champaign, IL, USA, from 2012 to 2013, Cornell University, Ithaca, NY, USA, from 2013 to 2014, and Rutgers University, Piscataway, NJ, USA, in 2014. Since 2014, he has been a Professor with the School of Information and Control, Nanjing University of Information Science and Technology, Nanjing, China. His research interests touch on the areas of pattern recognition and signal processing. He obtained the National Excellent Youth Fund in 2016 and was designated as the global Highly Cited Researchers in 2017. He is now a Senior Member of the IEEE.

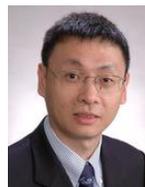
**Shuicheng Yan** (F'16-) received the Ph.D. degree from the Peking University in 2004. He is now the CTO of Yitu Tech, and also a (Dean's Chair) Associate Professor at the National University of Singapore. His research areas include computer vision, multimedia and machine learning, and he authored/co-authored more than 370 technical papers over a wide range of research topics, with the Google Scholar citations >58, 000 times and H-index-100. He is ISI Highly-cited Researcher of 2014, 2015 and 2016. He is/has been an Associate Editor of IEEE Trans. Knowledge and Data Engineering (TKDE), IEEE Trans. Circuits and Systems for Video Technology (TCSVT), ACM Trans. Intelligent Systems and Technology (TIST), and Journal of Computer Vision and Image Understanding. He is also a Fellow of the IEEE and the IAPR.

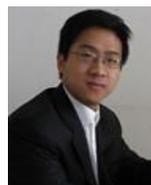
**Meng Wang** is a full professor in the Hefei University of Technology, China. He received the B.E. degree and Ph.D. degree in the Special Class for the Gifted Young and signal and information processing from the University of Science and Technology of China, Hefei, China, respectively. He previously worked as an associate researcher at Microsoft Research Asia, and then a core member in a startup in Bay area. After that, he worked as a senior research fellow in National University of Singapore. His current research interests include multimedia content analysis, search, mining, recommendation, and large-scale computing. He has authored 6 book chapters and over 100 journal and conference papers in these areas, including TMM, TNNLS, TCSVT, TIP, TOMCCAP, ACM MM, WWW, SIGIR, ICDM, etc. He received the paper awards from ACM MM 2009 (Best Paper Award), ACM MM 2010 (Best Paper Award), MMM 2010 (Best Paper Award), ICIMCS 2012 (Best Paper Award), ACM MM 2012 (Best Demo Award), ICDM 2014 (Best Student Paper Award), PCM 2015 (Best Paper Award), SIGIR 2015 (Best Paper Honorable Mention), IEEE TMM 2015 (Best Paper Honorable Mention), and IEEE TMM 2016 (Best Paper Honorable Mention). He is the recipient of ACM SIGMM Rising Star Award 2014. He is/has been an Associate Editor of IEEE Trans. on Knowledge and Data Engineering (TKDE), IEEE Trans. on Neural Networks and Learning Systems (TNNLS) and IEEE Trans. on Circuits and Systems for Video Technology (TCSVT). He is a senior member of IEEE.